	\documentclass[preprint,12pt,authoryear]{elsarticle}

	\usepackage{hyperref}
	\usepackage{amssymb}
	\usepackage{amsmath}
	\usepackage{amsmath,amsfonts}
	\usepackage{algorithmic}
	\usepackage{algorithm}
	\usepackage{array}
	\usepackage{tabularx}
	\usepackage[caption=false,font=normalsize,labelfont=sf,textfont=sf]{subfig}
	\usepackage{enumitem}
	\usepackage{textcomp}
	\usepackage{stfloats}
	\usepackage{url}
	\usepackage{verbatim}
	\usepackage{graphicx}
	\usepackage{cite}
	\usepackage{booktabs}
	\usepackage{multirow}
	\usepackage{siunitx}
        \usepackage[table,xcdraw]{xcolor}
        \definecolor{my_blue}{HTML}{00B0F0} 
        \definecolor{my_magenta}{HTML}{ED33D7}

\begin{document}

\begin{frontmatter}



	\title{COMO: Cross-Mamba Interaction and Offset-Guided Fusion for Multimodal Object Detection} 


	\author[label1]{Chang Liu}
        \author[label2]{Xin Ma} 
        \author[label3]{Xiaochen Yang}
        \author[label4]{Yuxiang Zhang} 
	\author[label1]{Yanni Dong\corref{mycorrespondingauthor}}
        \cortext[mycorrespondingauthor]{Corresponding author}
        \ead{dongyanni@whu.edu.cn}


	\affiliation[label1]{organization={School of Resource and Environmental Sciences}, 
		addressline={Wuhan University},
		city={Wuhan},
		postcode={430079},
		country={China}}
  	\affiliation[label2]{organization={State Key Laboratory of Information Engineering in Surveying, Mapping, and Remote Sensing}, 
		addressline={Wuhan University},
		city={Wuhan},
		postcode={430079},
		country={China}}
        \affiliation[label3]{organization={School of Mathematics and Statistics}, 
		addressline={University of Glasgow},
		city={Glasgow},
		postcode={G12 8QQ},
		country={UK}}
        \affiliation[label4]{organization={School of Geophysics and Geomatics}, 
		addressline={China University of Geoscience},
		city={Wuhan},
		postcode={430074},
		country={China}}

	\begin{abstract}
Single-modal object detection tasks often experience performance degradation when encountering diverse scenarios. In contrast, multimodal object detection tasks can offer more comprehensive information about object features by integrating data from various modalities. Current multimodal object detection methods generally use various fusion techniques, including conventional neural networks and transformer-based models, to implement feature fusion strategies and achieve complementary information. However, since multimodal images are captured by different sensors, there are often misalignments between them, making direct matching challenging. This misalignment hinders the ability to establish strong correlations for the same object across different modalities. In this paper, we propose a novel approach called the CrOss-Mamba interaction and Offset-guided fusion (COMO) framework for multimodal object detection tasks. The COMO framework employs the cross-mamba technique to formulate feature interaction equations, enabling multimodal serialized state computation. This results in interactive fusion outputs while reducing computational overhead and improving efficiency. Additionally, COMO leverages high-level features, which are less affected by misalignment, to facilitate interaction and transfer complementary information between modalities, addressing the positional offset challenges caused by variations in camera angles and capture times. Furthermore, COMO incorporates a global and local scanning mechanism in the cross-mamba module to capture features with local correlation, particularly in remote sensing images. To preserve low-level features, the offset-guided fusion mechanism ensures effective multiscale feature utilization, allowing the construction of a multiscale fusion data cube that enhances detection performance. The proposed COMO approach has been evaluated on three benchmark multimodal datasets consisting of RGB and infrared image pairs, demonstrating state-of-the-art performance in multimodal object detection tasks. It offers a solution tailored for remote sensing data, making it more applicable to real-world scenarios. The code will be available at \url{https://github.com/luluyuu/COMO}.

\end{abstract}

\begin{keyword}


    Object Detection \sep Multimodal Fusion \sep Mamba Model \sep Remote Sensing
\end{keyword}

\end{frontmatter}




\section{Introductions}

The object detection task enables rapid interpretation of images and the identification of object locations. As a key task in computer vision, it has been widely applied in various fields, including autonomous driving, remote sensing, and medical imaging \citep{zou2023survey,ma2024singlemodal}. However, in complex environments, such as low-light conditions, variable weather, and partial occlusion, the performance of single-modal object detection declines due to its limited capacity to effectively capture the salient features of the objects\citep{li2019illumination, CRmixing2024liu}.

Multimodal visual data, comprising images acquired from different sensors \citep{cao2019pedestriandetection,li2022deep} (e.g., RGB cameras, infrared sensors, Lidar, and Radar),  offers a richer set of feature attributes for object detection\citep{guan2019fusion, zhou2020improving}. By integrating multimodal data, complementary information can be leveraged, allowing the objects to exhibit distinct and prominent features across diverse scenarios.

Recent advancements in multimodal fusion techniques have led to significant improvements in detection performance. Approaches such as pixel-level fusion \citep{superyolo2023zhang}, feature-level fusion \citep{CFT2021qing,GMdetr2024xiao, zhang2019cross-modality}, and decision-level fusion \citep{zhu2023MFPT} enable the effective integration of data from multiple modalities. These methods exploit the complementary nature of multimodal data, maximizing the information available about the object, thereby improving detection accuracy and robustness in challenging environments.

\begin{figure}[t]
	\centering 
	\includegraphics[width=0.98\linewidth]{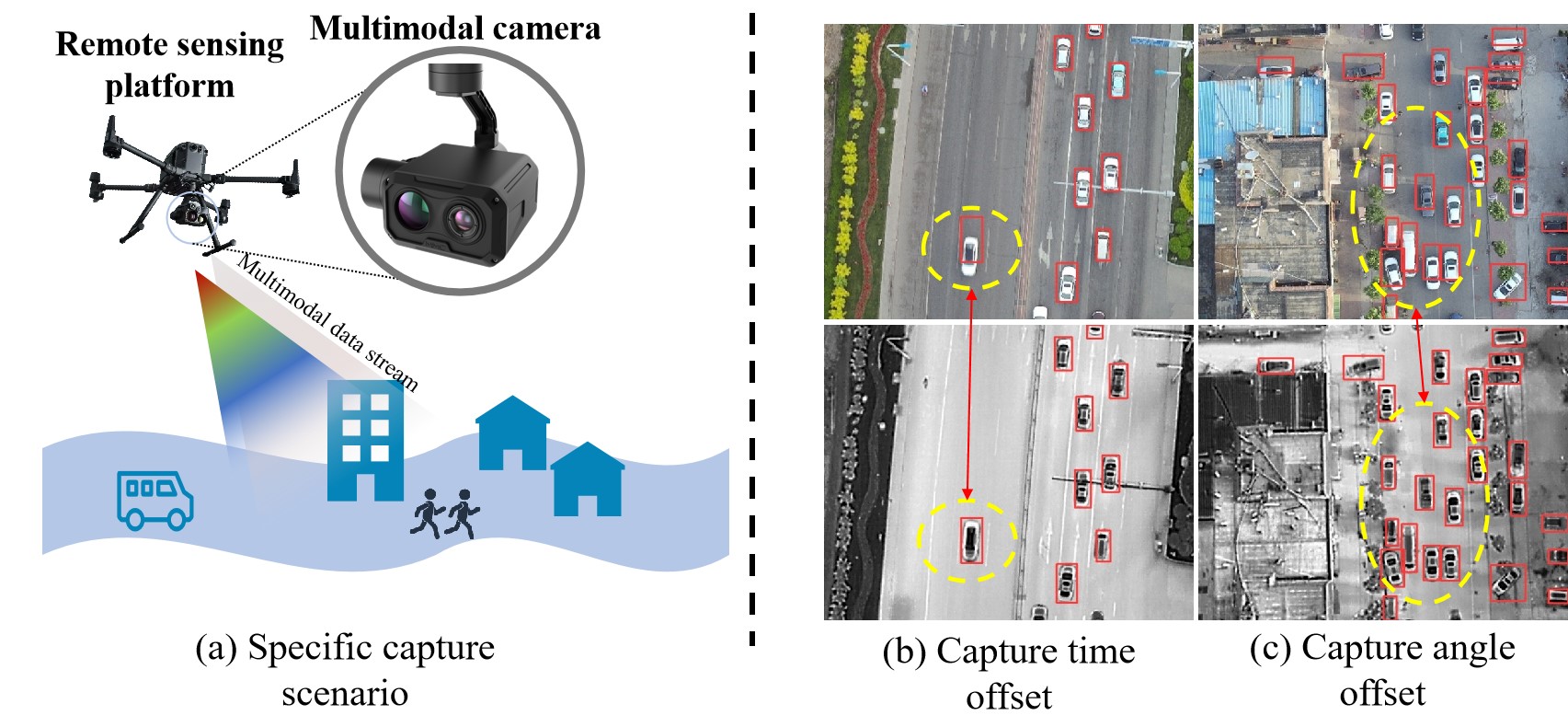}
	\caption{The phenomenon of offset in multimodal images. (a) Specific scenarios in multimodal data acquisition. (b) Offset due to differences in capture time. (c) Offset due to differences in capture angles.}
	\label{fig:offset}
\end{figure}
\begin{figure}[t]
	\centering
	\includegraphics[width=0.98\linewidth]{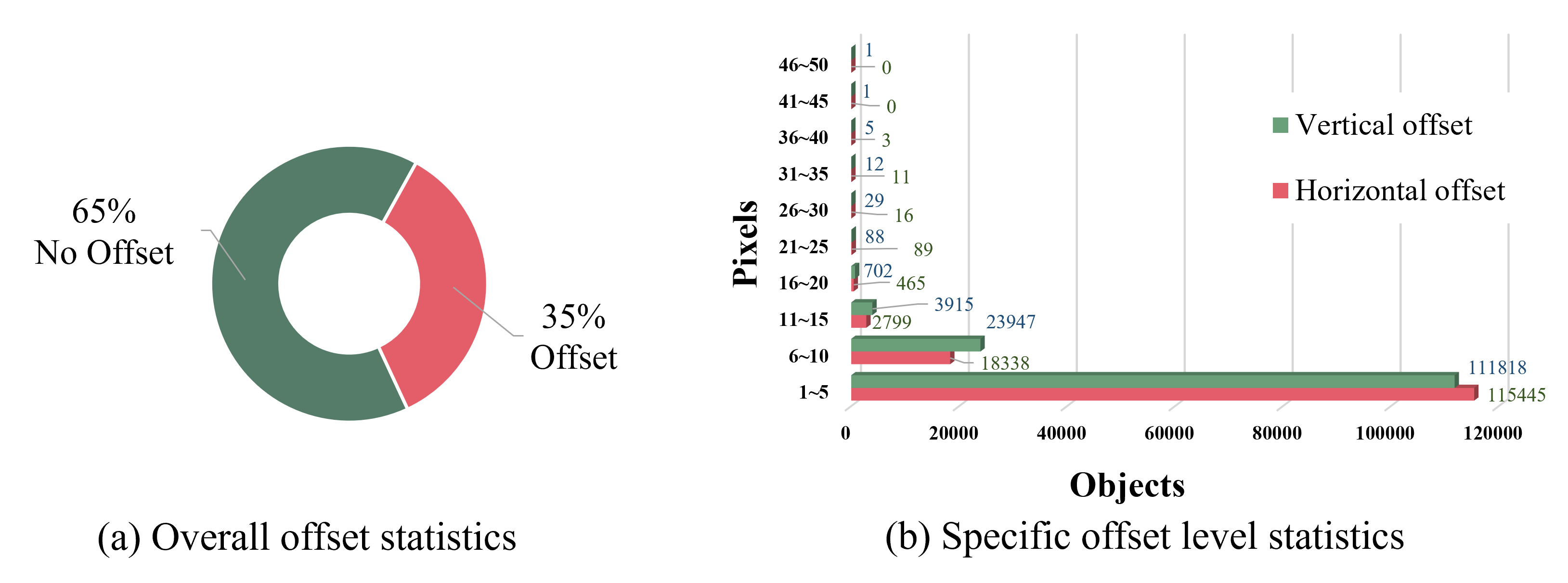}
	\caption{Offset statistics results using the DroneVehicle dataset as an example. (a) Overview of data offsets. (b) Specific offset level statistics.}
	\label{fig:offset_total}
\end{figure}

Despite these advances, there remain challenges in multimodal object detection tasks. One of those is the misalignment between data from different sensors \citep{song2024CMA}. This misalignment can be caused by variations in camera angles, capture times, or sensor characteristics, leading to discrepancies in object positions and features. As shown in Fig. \ref{fig:offset}, misalignment problems caused by differences in capture times and camera angles are common challenges in remote sensing multimodal data fusion tasks. Regarding the capture time issue, remote sensing data typically originate from airborne platforms, such as satellites and drones, which provide a high viewing angle and wide coverage area. When capturing a high-speed moving target, even a brief time interval between shots can cause significant positional changes due to the rapid movement, leading to noticeable offsets. This positional shift can compromise the accuracy of detection, particularly in applications that demand precise target recognition and tracking. As for the camera angle issue, in multimodal data acquisition, two or more cameras are commonly used for simultaneous capture, such as optical and infrared cameras. However, differences in camera positions and viewing angles often result in misaligned imaging positions of the same target across different modalities. These discrepancies complicate data alignment and fusion, potentially reducing application accuracy and hindering the establishment of strong correlations between objects across modalities, thereby making accurate object detection in multimodal data more challenging \citep{chen2024weakly}. More rigorously, we have analyzed the misalignment problem in DroneVehicle dataset, a large-scale UAV multimodal dataset \citep{Sun2022DroneVehicle}. The results reveal that up to 35\%  labels exhibit offset issues, with some labels showing significant displacement, as shown in Fig. \ref{fig:offset_total}. Objects with a pixel offset of 1 to 5 pixels accounted for over 90\% of all misaligned objects. This substantially impacts the accuracy of multimodal detection. Therefore, effective fusion strategies that account for offset corrections are crucial for enhancing the performance of multimodal remote sensing object detection tasks.

Furthermore,  multimodal data inherently contain more information compared to single-modal data, which increases the time required for data processing \citep{CFT2021qing}. Recently, feature-level fusion methods have gained popularity, yielding increasingly accurate results \citep{shen2024icafusion,GMdetr2024xiao}. However, the dual-branch feature extraction structures and multiscale fusion mechanisms employed in these methods significantly increase computational resource demands and processing time \citep{song2024CMA,zhu2023MFPT}. To address this issue, it is essential to develop efficient multimodal fusion strategies that can maintain high detection accuracy while streamlining the model for real-time processing.

In this work, to mitigate misalignment effects, reduce computational resources and time consumption, and enhance multimodal object detection performance, we propose a new method: CrOss-Mamba interaction and Offset-guided fusion (COMO) framework.  The COMO framework incorporates the novel mamba technique \citep{gu2023mamba} to develop the cross-mamba method, which formulates feature interaction equations, enabling serialized state computation. This approach reduces computational load and time consumption compared to current transformer-based methods.\citep{CFT2021qing,shen2024icafusion}. Additionally, COMO leverages high-level features, which are less prone to mismatches, to facilitate inter-modal interactions and information fusion, addressing positional offset issues arising from variations in camera angles and capture times.  COMO also incorporates a global and local scanning mechanism within the cross-Mamba method to capture features that encompass both global sequence information and local relevance, particularly in remote sensing images. To preserve low-level features, the offset-guided fusion mechanism ensures efficient multiscale feature utilization, thereby maximizing available information. Evaluated on three benchmark datasets, COMO demonstrates state-of-the-art performance in multimodal object detection tasks and offers a solution specifically tailored for remote sensing applications, enhancing its practical utility.

In summary, our contributions are threefold: 

\begin{itemize}
	\item We propose a multimodal object detection framework to address the offset issue in multimodal images. The framework employs the mamba interaction method to facilitate inter-modal information exchange and complementary fusion. Additionally, it integrates a global and local scanning mechanism to capture both global and local correlation features.
	\item We design an offset-guided fusion method to address the issue of low-level feature loss that arises from relying solely on high-level features for interaction. This approach allows high-level features to guide the fusion of low-level features, thereby maximizing information retention and minimizing the impact of offsets.
	\item We conduct experiments on three benchmark datasets with varying perspectives and compare our method against several related approaches. The results show that our proposed method achieves optimal performance across diverse scenarios. Furthermore, we meticulously examined the impact of the model components, confirming that our method effectively meets real-world application requirements.
\end{itemize}

\section{Related works}

In recent years, research in visual multimodal fusion and object detection has gained significant attention due to the limitations of single-modal approaches in complex environments. This section reviews key contributions to the fields of visual multimodal fusion, multimodal object detection, and a notable deep learning model, the mamba model.

\subsection{Visual Multimodal Fusion}
 Single-modal data is highly susceptible to situational changes, often leading to poor detection results \citep{yolors2020Sharma, CRmixing2024liu}. For instance, RGB images may perform well in clear conditions, but their effectiveness is significantly diminished in complex scenarios such as nighttime or cloudy weather \citep{yolo2016Redmon, DETR2020carion}. Introducing additional visual modalities can greatly enhance the robustness of vision tasks by compensating for these limitations \citep{superyolo2023zhang, GMdetr2024xiao}.

Visual multimodal fusion leverages data from multiple sensors or modalities (e.g., RGB, infrared) to enrich feature representations of objects. Various fusion methods have been proposed to exploit complementary information across modalities. Early approaches can be categorized into two groups: transform domain-based methods and spatial domain-based methods. Transform domain fusion methods are the focus of early research, and typical methods \citep{yin2018transformfusion} include wavelet transform, curved wavelet transform, Laplace pyramid. These methods can effectively retain the detailed information in multimodal images by decomposing and reconstructing the image signals at different scales and frequencies. \citet{Pixel2017li} reviewed the development of pixel-level fusion techniques, pointing out the wide application of transform techniques such as wavelets and curvilinear waves in image fusion. With the development of research, multimodal image fusion methods combining wavelet transform and deep learning have also emerged. \citet{deng2020deng} proposed a fusion method combining wavelet transform which showed better performance. Spatial domain-based image fusion methods directly process the original image pixel values and utilize local features, spatial frequency, or gradient information for fusion \citep{meher2019spatialfusion}. These methods decompose and reconstruct the image by means of Laplace pyramid,  Non-Subsampled Contour Wave Transform (NSCT), etc., and are more suitable for dealing with edge and detail information in the image. For example, the Laplace pyramid proposed by \citet{laplacian1987burt} became a classical method for multimodal image fusion, which was later widely used in remote sensing, medicine, and other image processing fields. \citet{nejati2015multi} proposed a new multifocal image fusion method based on convolutional sparse representation, which achieved good results in multifocal fusion applications.

In recent years, with the development of deep learning technology, deep learning models such as convolutional neural networks (CNNs) and generative adversarial networks (GANs) have gradually been applied in multimodal image fusion. Deep learning methods can automatically learn image features and perform end-to-end fusion processing, which significantly improves the accuracy and efficiency of fusion. \citet{image2016liu} introduced CNN into infrared and visible image fusion for the first time, demonstrating the potential of deep learning methods in fusion tasks. \citet{zhang2023visible} further investigated a deep learning-based method for infrared and visible image fusion, which realized high-quality image fusion in complex environments.

In addition, the introduction of Generative Adversarial Networks (GAN) also provides new ideas for image fusion. The GAN model proposed by \citet{goodfellow2014gan} performs well in image generation and reconstruction. \citet{rao2023tgfuse} successfully realized multimodal image fusion using GAN, which shows better results than the traditional methods. The introduction of deep learning techniques has brought great innovations in the field of image fusion, but the long training time, the need for large amounts of data, and the poor interpretability of the model are still the shortcomings and points for future focus of research.

\subsection{Multimodal Object Detection}
Unlike multimodal image fusion which aims at obtaining better visualization, multimodal object detection task is result-oriented and is more task-specific. Visual multimodal object detection extends traditional object detection tasks by incorporating multimodal data to enhance detection performance. It aims to utilize the complementary information between different modalities to enhance the robustness and accuracy of object detection, especially in complex environments, bad weather, or occlusion situations. Depending on the fusion strategy, multimodal object detection can be categorized into pixel-level fusion, feature-level fusion, and decision-level fusion methods.

Pixel-level fusion directly splices or overlays raw data from different modalities (e.g., RGB images and infrared images) and inputs them into the same object detection network. This approach often does not distinguish between modalities, but rather unifies all data as input to the network. Since the effect of direct splicing and then detection is not ideal, YOLOrs \citep{yolors2020Sharma} proposes a two-stage fusion method, which involves using deep convolutional networks to extract features from each modality independently. The extracted features are then fused through concatenation and element-wise cross-product operations, maximizing the information in the fused data body. SuperYOLO \citep{superyolo2023zhang} proposes a fusion method called multimodal fusion(MF) to extract complementary information from various data to improve the small target detection task in remote sensing, and it pioneers the super-resolution branch to enhance the accuracy of the backbone feature extraction network.

The feature-level fusion approach is characterized by multiscale feature fusion and richer information retention, which enhances model robustness, strengthens generalization ability, and reduces information loss, making it particularly popular in current research \citep{guo2020augfpn}. Numerous methods \citep{CFT2021qing, shen2024icafusion, GMdetr2024xiao, song2024CMA} have been proposed to improve the multimodal object detection results continuously. CFT \citep{CFT2021qing} pioneered the use of the transformer framework in the field of multimodal object detection, which is based on the principle of splicing multimodal data patches and feeding them simultaneously into a self-attention structure to obtain inter-modal global attention results. ICAFusion \citep{shen2024icafusion}, on the other hand, utilizes the cross-attention mechanism for inter-modal feature interaction fusion. CMADet \citep{song2024CMA}  aims to solve the problem of data misalignment between two modalities and realize multiscale feature alignment and detection. GM-DETR \citep{GMdetr2024xiao} utilizes the state-of-the-art RT-DETR framework and proposes a novel training strategy, namely the modal complementation strategy. Performing two-stage training can give the model a better modal adaptation effect.  OAFA \citep{chen2024weakly} is another multimodal detection method that accounts for feature misalignment between modalities. Its approach focuses on mitigating the impact of modality gaps on multimodal spatial matching by obtaining modality-invariant features within a shared subspace, thereby estimating precise offset values. 

In decision-level fusion methods, object detectors are first independently trained for each modality. The results from each detector are then combined using strategies such as voting, weighted averaging, and other techniques to derive the final detection outcome. The MFPT method \citep{zhu2023MFPT} employs intra-modal and inter-modal transforms to enhance individual modal features, ultimately realizing an enhanced, feature-based decision-level fusion approach.

The multimodal data are acquired by two or more sensors, which are side-by-side and acquire data with different field-of-view angles. Moreover, there are data acquisition time differences between different sensors, which have a huge impact on objects moving at high speeds. Therefore, it is necessary to take into account the offset problem between the data when data fusion is performed. Meanwhile, multimodal object detection has more multi-branch feature extraction networks and feature fusion modules compared to single-modal object detection, so the time consumption increases. In order to make the multimodal object detection model have real-time capability, it is necessary to improve the detection accuracy while streamlining the model.

\subsection{Mamba model}

The mamba model \citep{gu2023mamba} is an efficient sequence feature extraction model that has emerged in recent times. Its core idea is to selectively use the state spaces model (SSM) in sequence modeling to balance modeling power and computational efficiency. Compared to traditional recurrent neural networks (RNNs) or autoregressive models (e.g., Transformer), mamba maintains linear time complexity when processing long sequences by utilizing an efficient state space model. As a result, it can be computed efficiently even as the sequence length increases. When applied to the field of computer vision, mamba has achieved excellent results across various tasks.

Vision mamba \citep{zhu2024vim} is the first approach to introduce mamba models from natural language into computer vision. It draws on the ideas of ViT \citep{dosovitskiy2020vit} and proposes a bi-directional scanning mechanism to serialize the image data, thus allowing the overall model to achieve global attention and feature association. It proves the effectiveness of the mamba model for a wide range of visual tasks, opening up new paths for the field of computer vision. ChangeMamba \citep{chen2024changemamba} explores for the first time the potential of the mamba architecture for remote sensing change detection tasks. U-Mamba \citep{ma2024umamba} designs a hybrid CNN-SSM module that integrates the local feature extraction capability of convolutional layers with the ability of SSM to capture long-range dependencies for use in the field of medical image segmentation. FusionMamba \citep{xie2024fusionmamba} explores the potential of the SSM model in the field of image fusion, which utilizes the mamba model to design a U-Net structure to fuse data from both modalities. LocalMamba \citep{huang2024localmamba} introduces a novel local scanning strategy that divides the image into different windows to efficiently capture local dependencies while maintaining a global perspective.

\section{Methods}
\begin{figure}[t]
	\centering
	\includegraphics[width=1.01\linewidth]{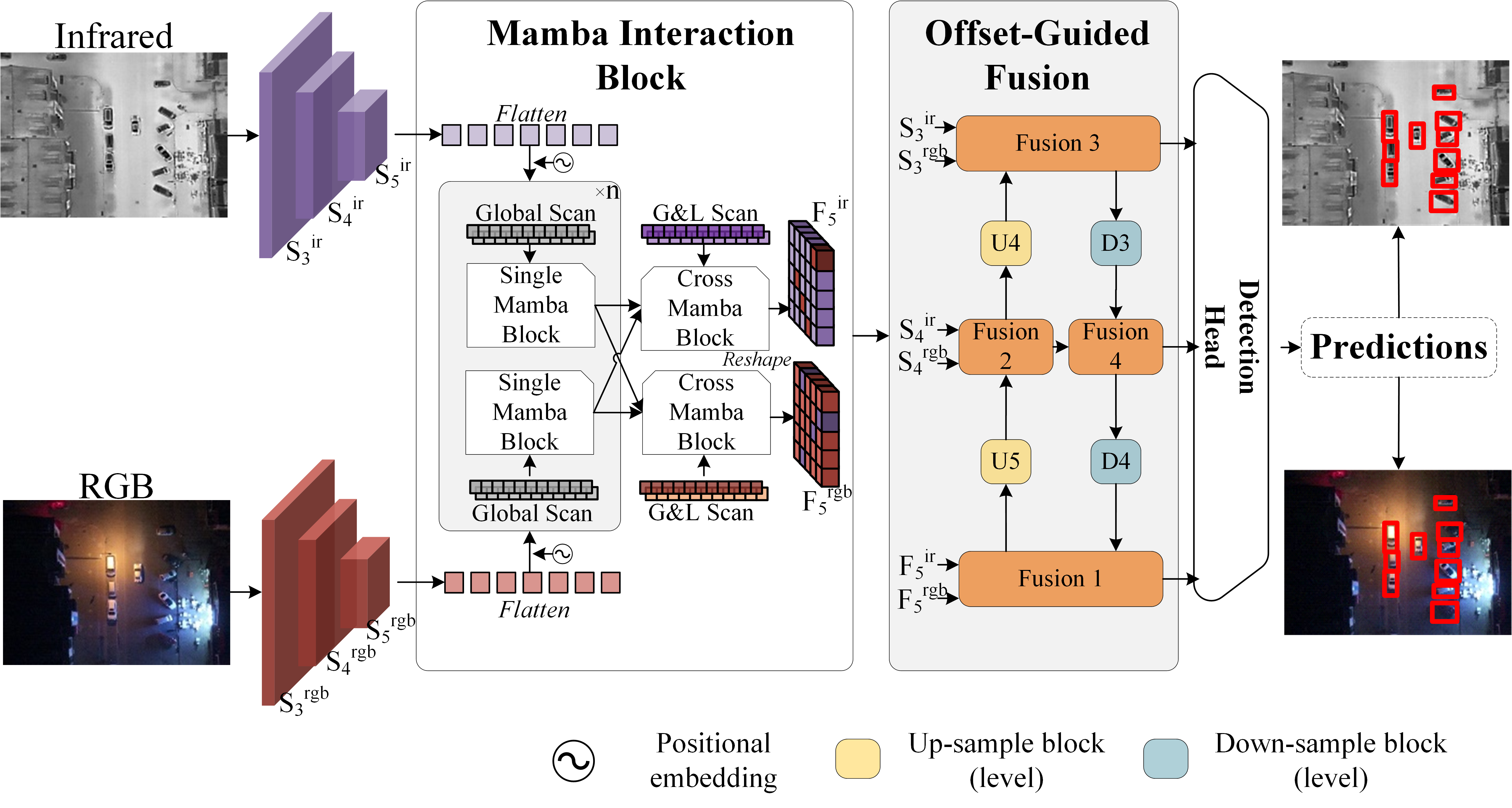}
	\caption{Architecture of COMO framework. The framework consists of three main components: Mamba Interaction Block, Global and Local Scan method, and Offset-Guided Fusion. The Mamba Interaction Block is used to extract high-level features and perform inter-modal interaction. The Global and Local Scan method is used to strengthen the local feature association. The Offset-Guided Fusion module is used to fuse high-level features and low-level features.} 
	\label{fig:framework}
\end{figure}

This section presents a detailed overview of the COMO approach, as illustrated in Fig. \ref{fig:framework}. We first introduce the overall structure of the COMO approach, followed by detailed descriptions of the key components: Mamba Interaction Block, Global and Local Scan Method, and Offset-Guided Fusion.

\subsection{Overall Structure}
Given a pair of visible and infrared images \{$x_{rgb}$, $x_{ir}$\}, the proposed COMO approach obtains detection results beyond a single modality by performing intermodal interaction and fusion. To be specific, $x_{rgb}$ and $x_{ir}$ are passed through two CNN backbones with the same structure, which allows for the extraction of salient features specific to each modality, then obtaining multiscale feature extractions \{$S_3^{ir}$,$S_4^{ir}$,$S_5^{ir}$,$S_3^{rgb}$,$S_4^{rgb}$,$S_5^{rgb}$\} (the feature maps from stages 3, 4, and 5 of the backbone feature extraction network.). To minimize the effect of offset on fusion, only the highest-level features \{$S_5^{ir}$,$S_5^{rgb}$\} extracted by the CNN network were selected for the mamba interaction block. The multiscale feature extraction results \{$S_3^{ir}$,$S_4^{ir}$,$S_3^{rgb}$,$S_4^{rgb}$\}are fed into the offset-guided fusion network along with the high-level interaction features \{$F_5^{ir}$,$F_5^{rgb}$\}to achieve unbiased fusion. The final result is produced by the detection head after the final offset-guided fusion network.

To provide the model with an advantage in both real-time performance and accuracy, it is crucial to carefully handle offsets. Unlike previous approaches, we choose to use the highest-level features as interaction features. This decision is based on the fact that high-level data contains semantic information about objects, where the maximum offset within their spatial receptive field has less impact compared to lower-level features. This relationship can be explained by the following equation:

\begin{equation}
	A_{\text {intersection }}=|w_{blk}-\Delta x| \times |h_{blk}-\Delta y|.
\end{equation}
Here, \(\Delta x\) and \(\Delta y\) are offsets, both of which are fixed for a multimodal image pair. The \(w_{blk}\) and \(h_{blk}\) are the width and height of image blocks with different levels of downsampling. Since the offsets are fixed, larger \(w_{blk}\) and \(h_{blk}\) are needed to be able to obtain a larger area of intersection \(A_{\text {intersection }}\). At the same time, using only high-level features for intermodal interaction can significantly reduce the amount of computation and improve the real-time performance of the model.

To retain the lower-level features and avoid duplicating the fusion structure with the object detection neck, we design the Offset-Guided Fusion method. This method reduces both computation and processing time. Simultaneously, using high-level features, which are less influenced by offsets, as a guide to bridge modalities helps mitigate the impact of offsets on low-level features, thereby ensuring the efficient utilization of these features.

We implement the proposed methods in YOLOv5 to enable comparisons with other approaches within the same framework, and in YOLOv8 to achieve enhancements based on the new baseline framework.

\subsection{Mamba Interaction Block}

\begin{figure}[!t]
	\centering
	\includegraphics[width=0.9\linewidth]{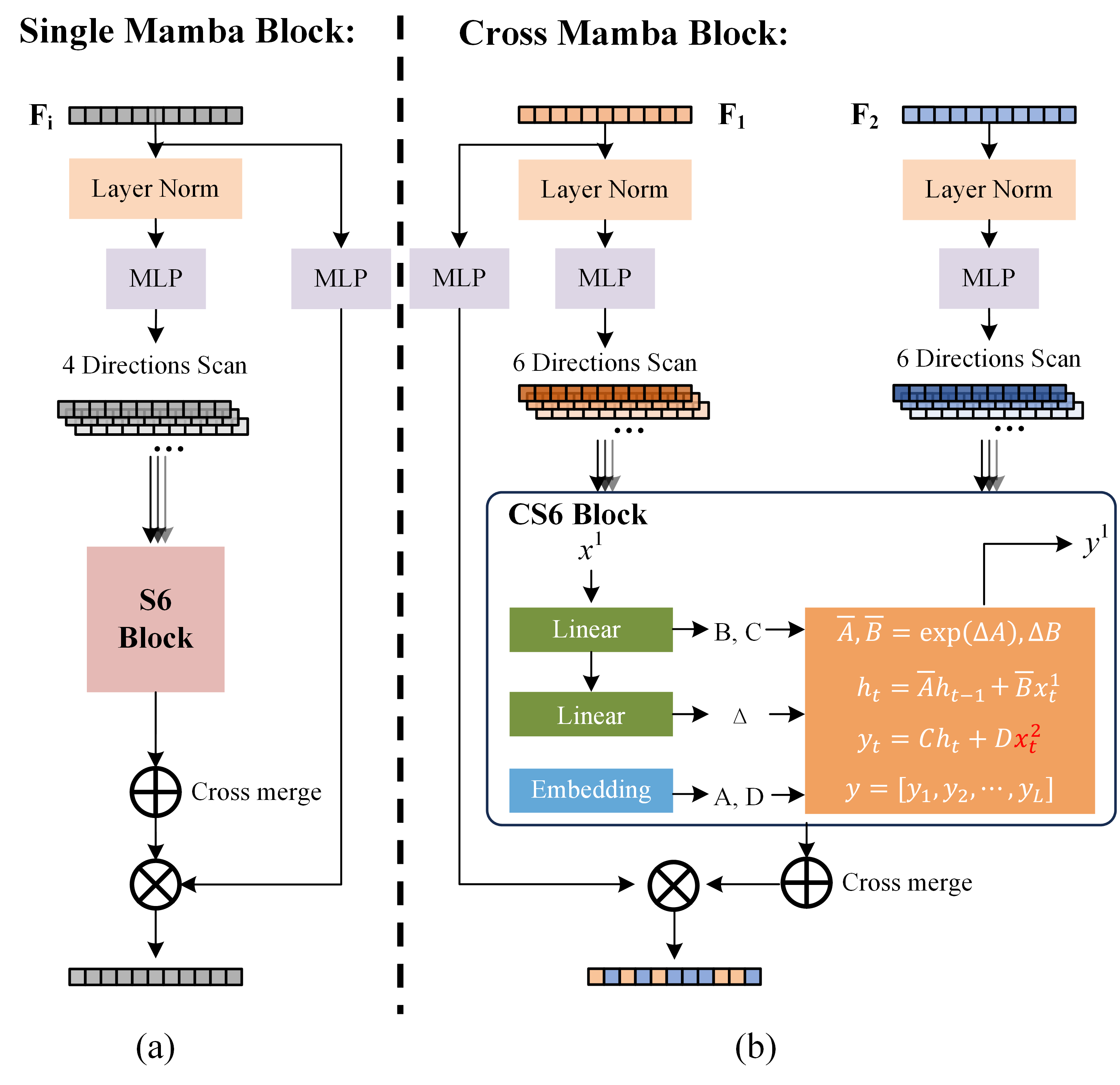}
	\caption{Mamba interaction block. The block consists of two modules: (a) single-mamba block and (b) cross-mamba block. The single-mamba block is used to extract features from single-modal data, while the cross-mamba block is used to interact between multimodal data.}
	\label{fig:crossmamba}
\end{figure}

The Mamba Interaction Block is shown in Fig. \ref{fig:crossmamba}. It comprises two modules, the single-mamba block and the cross-mamba block. To leverage mamba's efficient feature extraction, we first implement two single-mamba blocks to extract features from single-modal data. These blocks serialize the output from the CNN backbone network, capturing global historical state information through various scanning modes. The blocks adapt the extracted features into sequences by repeatedly applying these operations. This method is specifically used for high-level features {$S_5^{ir}$ and $S_5^{rgb}$}, which contain substantial semantic information and are minimally affected by spatial offsets.

For the input features represented as $S_{in}$, we apply adaptive max pooling and mean pooling to construct feature matrices $F_{in}\in\mathbb{R}^{H\times W \times C}$, ensuring consistent dimensions across varying image sizes:
\begin{equation}
	F_{in} = \mathcal{P}_{avg}(S_{in}) + \mathcal{P}_{max}(S_{in}).
\end{equation}

We then perform a deep feature mapping of $F_{in}$ and add the dropout operator \citep{srivastava2014dropout} thus making the model adaptive.
\begin{equation}
	F_{m} = \textbf{Drop}(\mathcal{F}^{h\rightarrow C}(\textbf{Silu}(\mathcal{F}^{C\rightarrow h}(F_{in})))).
\end{equation}
Here, $h$ is the channels of hidden features in the mapping process, $\mathcal{F}(\cdot)$ is the linear mapping operation, $\textbf{Drop}(\cdot)$ means the randomized discarding of neurons with some probability, and $\textbf{Silu}(\cdot)$ means the  activation function for nonlinearization.
The resulting tensor $F_m$ is then flattened into a token sequence, $I_{in}\in\mathbb{R}^{H W \times C}$, simulating sequential data for state-space model algorithms. To mitigate the loss of two-level spatial information, we incorporate a learnable positional embedding $P \in\mathbb{R}^{H W \times C}$ which provides explicit positional encoding. Finally, we establish shortcut data streams for manipulation, maintaining the integrity of the original feature extraction. After that, $I_{in}$ will be scanned in four directions expanding the serialization approach with positional encoding, thus expanding the data distribution. Then, the scanning results of each direction will go through S6 block separately for sequence feature state space model feature extraction, resulting in four outputs denoted as $y_i$:
\begin{equation}
	\left\{
	\begin{aligned}
		x_{i}   & = \textbf{cross scan}_{i}(I_{in}),            \\
		y_{i}   & = \textbf{S6}_{i}(x_{i}),                     \\
		I_{out} & = \sum_{i=1}^4\textbf{reverse scan}_{i}(y_i).
	\end{aligned}
	\right.
	\quad i=1,2,3,4.
\end{equation}
Here, $\textbf{cross scan}(\cdot)$ indicates a four-direction scanning method, as shown in Fig. \ref{fig:crossmamba}. $\textbf{S6}(\cdot)$ is the state-space model (SSM) structure of the mamba model. The $\textbf{reverse scan}(\cdot)$ represents that $y_i$ obtained after feature extraction needs to go through the reverse scanning process of $\textbf{cross scan}(\cdot)$ to get its feature expression under the original sequence structure.

The S6 block represents an enhancement of the SSM model, which as a continuous system can map 1D inputs $x(t)\in \mathbb{R}$ to outputs $y(t)\in \mathbb{R}$ carrying historical states $h(t)\in \mathbb{R}^N$ via hidden state space equations:
\begin{equation}
	\left\{
	\begin{aligned}h^{\prime}(t)&=\mathbf{A}h(t)+\mathbf{B}x(t),\\y(t)&=\mathbf{C}h(t) + \mathbf{D}.\end{aligned}
	\right.
\end{equation}
Here, \textbf{A} denotes the evolution parameter, $\textbf{B}$ and $\textbf{C}$ are the projection parameters, and $\textbf{D}$ is the skip connection. Because the historical state influences the output of the SSM model, it is powerful in the processing of sequential data. 

When the SSM is used in the deep learning field, the sequence data need to be discretized. \citet{gu2023mamba} introduced a timescale parameter, denoted as $\Delta\in\mathbb{R}$ to transform the continuous parameters $\textbf{A}$ and $\textbf{B}$ into discrete as $\overline{\textbf{A}}$ and $\overline{\textbf{B}}$. By employing the zero-order hold (ZOH) \citep{schreier2010zeroorder} as the transformation algorithm, the discrete parameters are expressed as follows:
\begin{equation}
	\left\{
	\begin{aligned}&\overline{\mathbf{A}}=\exp(\Delta\mathbf{A}),\\&\overline{\mathbf{B}}=(\Delta\mathbf{A})^{-1}(\exp(\Delta\mathbf{A})-\mathbf{I})\cdot\Delta\mathbf{B}\approx\Delta\mathbf{B}.\end{aligned}
	\right.
\end{equation}
Here, $\overline{\mathbf{A}}\in \mathbb{R}^{N\times N}$, while $\overline{\mathbf{B}}\in\mathbb{R}^{N\times 1}$. After that, the discretized state space equation can be expressed as:
\begin{equation}
	\left\{
	\begin{aligned}&h_t=\overline{\mathbf{A}}h_{t-1}+\overline{\mathbf{B}}x_t,\\&y_t=\mathbf{C}h_t+\mathbf{D}x_t.\end{aligned}
	\right.
\end{equation}
Here, the $x_t$ represents the discretized input data, not continuous functions $x(t)$, $\mathbf{C}\in\mathbb{R}^{1\times N}$ and $\textbf{D}\in\mathbb{R}^1$, $y_t$ is the output of this state, and the final output is the set of results for all states:
\begin{equation}
	\label{concat-out}
	\textbf{Y}_s=[y_1, y_2, \ldots ,y_L].
\end{equation}
Here, $L$ is the sequence length, which is equal to \(H\times W\), and \(\textbf{Y}_s\) is the output of a single-mamba block. We constructed $n$ single-mamba blocks of the same structure to deeply extract state space features.

Inspired by the fusion-mamba architecture \citep{xie2024fusionmamba}, we developed the cross-mamba block to facilitate feature interaction between multimodal data. Unlike the single-mamba block, which operates on single-modal input, the cross-mamba block uses multimodal inputs as the foundation for feature interactions.
The calculation process is:
\begin{equation}
	\left\{
	\begin{aligned}
		x_{i}^1, x_{i}^2 & = \textbf{cross scan}_{i}(F_s^1, F_s^2),      \\
		y_{i}            & = \textbf{CS6}_{i}(x_{i}^1, x_{i}^2),         \\
		I_{out}          & = \sum_{i=1}^4\textbf{reverse scan}_{i}(y_i).
	\end{aligned}
	\right.
	\quad i=1,2,\ldots,6.
\end{equation}
Here, $F_s^1, F_s^2$ are the multimodal inputs for the cross-mamba block, and $\textbf{CS6}$ is the core computational method of the cross-mamba block, which is specified:
\begin{equation}
	\left\{
	\begin{aligned}
		h_t & = \overline{\mathbf{A}}h_{t-1} + \overline{\mathbf{B}}x_t^1, \\
		y_t & = \mathbf{C}h_t + Dx_t^2.
	\end{aligned}
	\right.
\end{equation}
Here, the $x_t^1, x_t^2$ indicate serialized state inputs for two modal data.  The core idea is to treat the input from the first modality as the historical state, using it to interact with the input from the second modality. This interaction generates deeply interconnected cross-modal data, enabling the construction of a complementary data structure. Then the outputs of the two CS6 blocks are spliced according to Eq. \ref{concat-out} to obtain the final output as $\mathbf{F}_{5}^{rgb}$ and $\mathbf{F}_5^{ir}$.

\subsection{Global and Local Scan Method}
\begin{figure}[t]
	\centering
	\includegraphics[width=0.75\linewidth]{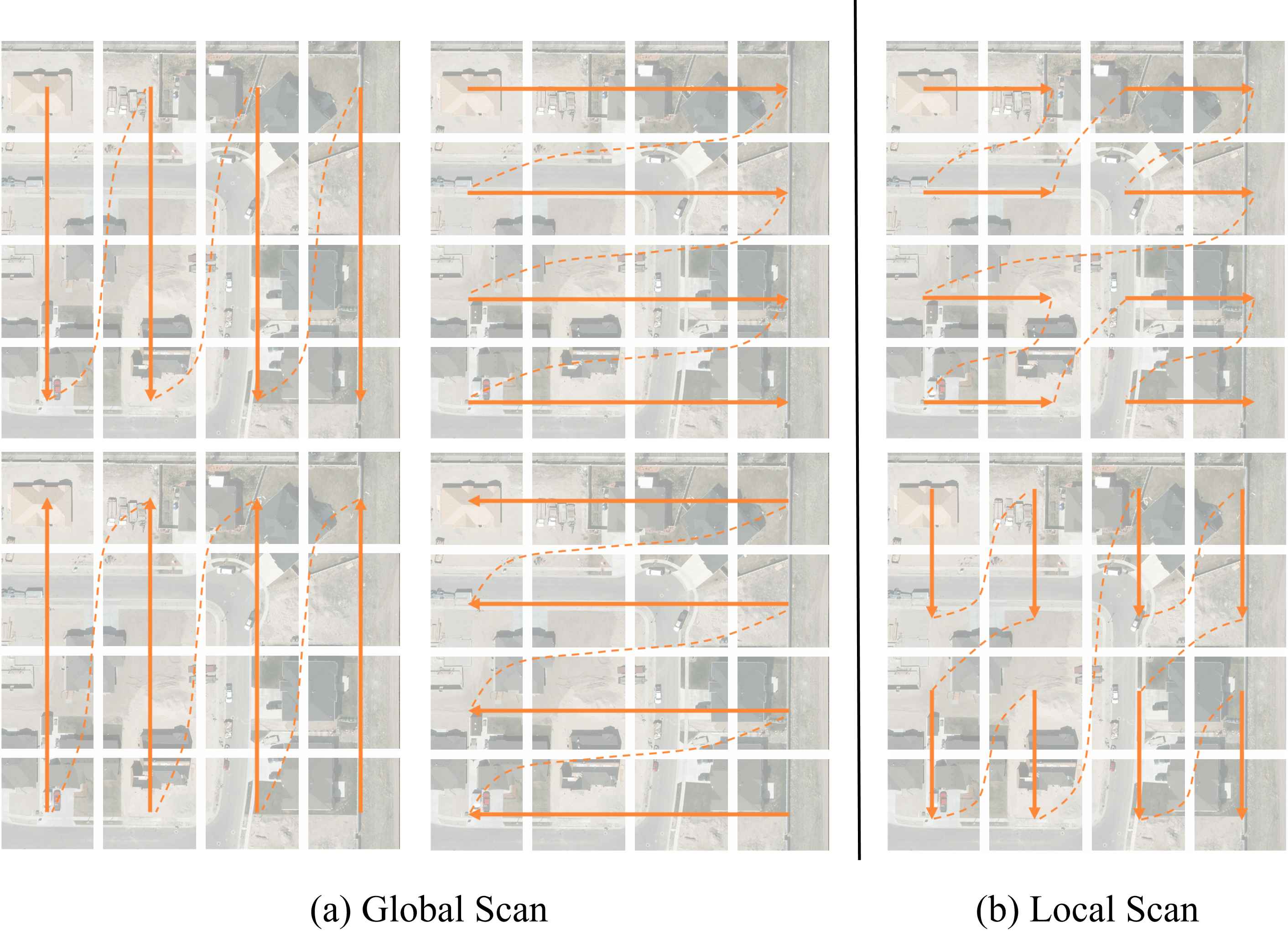}
	\caption{Different scanning mechanisms. (a) Global scanning. (b) Local scanning.}
	\label{fig:local-scan}
\end{figure}
The core of the mamba model is the S6 block, which excels at processing one-dimensional causal sequential data. However, the typical method of serializing images in visual imagery often relies on a global sequential scanning approach, similar to the global sequential modeling in Vim \citep{zhu2024vim} and VMamba \citep{liu2024vmamba}. While this approach is effective for language modeling, where understanding dependencies between consecutive words is essential, it does not align with the non-causal nature of 2D spatial relationships in images. Simple global serialized scanning can weaken the model's ability to discern these spatial relationships effectively.

Unlike transformers, which compute relationships between all spatial locations, the mamba model focuses on the state relationships between neighboring locations. In remote sensing imagery, the relationships between objects and the global context are often less critical than those within the visual images. Consequently, the use of global scanning can undermine the strengths of the mamba model by weakening the association of objects distributed in a local region. 

To address this issue, we propose the local scan method (LS) that divides the image into different windows to capture local dependencies while maintaining a global perspective efficiently. This strategy allows the model to focus on local relationships within each window while still considering the global context. By incorporating local scanning into the mamba model, we aim to enhance the model's ability to capture spatial relationships in visual imagery, particularly in remote sensing applications. As shown in Fig. \ref{fig:local-scan}, the LS method divides the image into multiple windows and scans each window sequentially. The local window size is a hyperparameter that can be adjusted based on the specific task requirements. We set the window size at most one-third of the image size to ensure that the model captures local dependencies effectively. As Fig. \ref{fig:crossmamba}(b), Cross Mamba Block part shows we add 2 directions of local scan to the cross-mamba block to build the Global and Local Scan method which enabling the mamba interaction block to capture both local and global spatial relationships, enhancing its performance in visual multimodal object detection tasks.

\subsection{Offset-Guided Fusion}

To address the limitation of high-level features, which are less affected by offset but lack low-level texture details, we design an Offset-Guided Fusion module. This module integrates high-level features after interaction with low-level features via a top-down feature pyramid networks (FPN) \citep{lin2017FPN} and a bottom-up path aggregation network (PAN) \citep{li2018pan}. This process allows the high-level features to guide the low-level features, mitigating the offset problem while preserving the low-level information. At the same time, it combines the fusion module with the object detection neck module to avoid structural duplication and thus reduce the number of parameters and computational time. The module operates through two branches: the top-down FPN and the bottom-up PAN.

\begin{figure}[t]
	\centering
	\includegraphics[width=0.7\linewidth]{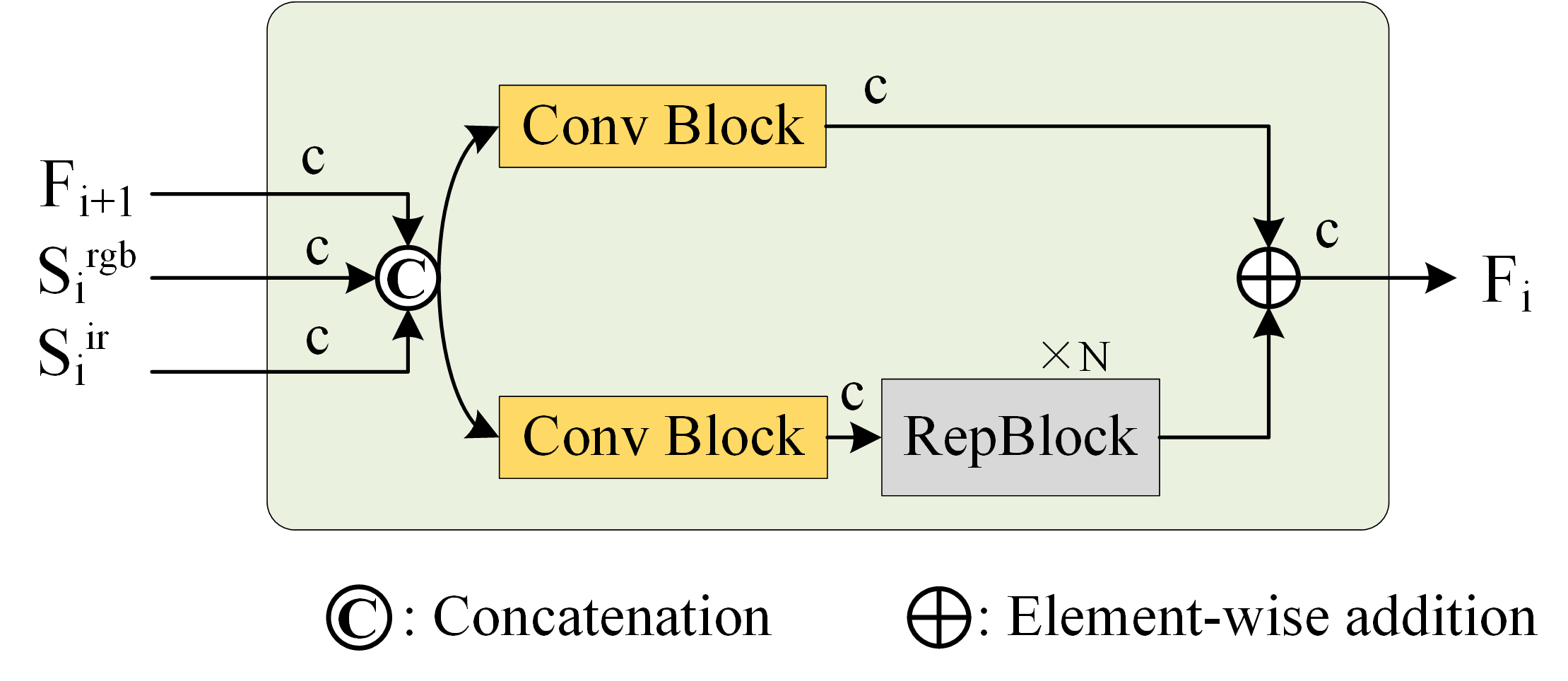}
	\caption{Offset-guided fusion method. It leverages high-level features to guide the fusion of low-level multimodal data, thereby mitigating the impact of biases. $F_{i+1}$ represents the higher level feature, $S_i^{rgb}$ and $S_i^{ir}$ represent the lower level features. The variable c denotes the number of channels in the feature maps.}
	\label{fig:fusion}
\end{figure}

The Offset-Guided Fusion method is a multi-level fusion module, as shown in the Fig. \ref{fig:fusion}. It utilizes high-level features without offset to guide the fusion of low-level features across multiple scales. Specifically, as illustrated in the figure, the fusion structure receives three types of input data: high-level features and low-level features from two different modalities. By implementing channel reconstruction and channel residual preservation, this approach builds a dual-branch feature fusion model, maximizing the information flow and achieving offset-guided fusion. The detailed process is as follows:
\begin{equation}
	\mathcal{F}(\mathbf{x}) = \sum_{i=1}^{N} \left( \text{ConvBlock}_i(\mathbf{x}) + \text{RepBlock}(\text{ConvBlock}_i(\mathbf{x})) \right).
\end{equation}
Here, $\mathbf{x}$ is the input feature after  concatenation, $\text{ConvBlock}_i(\cdot)$ is the convolutional channel residual preservation block, and $\text{RepBlock}(\cdot)$ is the channel reconstruction block. The fusion process occurs across multiple scales, where high-level features guide the fusion of low-level features. This approach effectively mitigates the offset issue while preserving low-level texture details, thereby enhancing the model’s performance in multimodal object detection tasks.

\section{Experiments}

\begin{table}[t]
	\centering
	\caption{Dataset summary statistics \label{tab:dataset-statistics} }
	\begin{tabular}{ccccc}
		\toprule
		Datasets                       & Split & Images & Annotation & Resolution                          \\
		\midrule
		\multirow{2}{*}{DroneVehicle} & Train & 17980  & 316,411    & \multirow{2}{*}{\(640\times512\)}   \\
		                              & Test  & 3463   & 24,490     &                                     \\
		\midrule
		\multirow{2}{*}{LLVIP}        & Train & 12025  & 34135      & \multirow{2}{*}{\(1280\times1024\)} \\
		                              & Test  & 3463   & 8302       &                                     \\
		\midrule
		\multirow{2}{*}{VEDAI}        & Train & 1089   & 3276       & \multirow{2}{*}{\(512\times512\)}   \\
		                              & Test  & 121    & 364        &                                     \\
		\bottomrule
	\end{tabular}
\end{table}
We present the experimental settings and results to validate the effectiveness of the COMO approach in multimodal object detection tasks. The experimental results demonstrate the effectiveness of the COMO approach in achieving state-of-the-art performance in multimodal object detection tasks.

\subsection{Experimental settings}
To comprehensively compare the performance of the models, we selected three datasets, each offering a different perspective, as benchmarks: DroneVehicle \citep{Sun2022DroneVehicle}, LLVIP \citep{jia2021llvip}, and VEDAI \citep{razakarivony2016vedai}. Detailed statistics for each dataset are presented in Table \ref{tab:dataset-statistics}. For the comparison algorithms, we selected several highly relevant methods and reproduced them exactly to obtain comparable results. These methods include YOLOrs \citep{yolors2020Sharma}, CFT \citep{CFT2021qing}, SuperYOLO \citep{superyolo2023zhang}, GHOST \citep{zhang2023ghost}, MFPT \citep{zhu2023MFPT}, ICAFusion \citep{shen2024icafusion}, GM-DETR \citep{GMdetr2024xiao}, DaFF \citep{althoupety2024daff}, and CMADet \citep{song2024CMA}.

We implement the COMO approach using two baseline object detectors YOLOv5 and YOLOv8. We used an NVIDIA RTX3090 GPU for all of our experiments. The size of the training data and test data was set to \(640\times640\) pixels in all experiments. For the large-scale DroneVehicle and LLVIP datasets, we use the smaller YOLOv5s and YOLOv8s model architectures as benchmarks, setting the training epochs to 150 to reduce training time and resource consumption. For the smaller VEDAI dataset, we select the larger YOLOv5l models, increasing the training epochs to 300 to maximize accuracy. In order to obtain more accurate experimental results and to speed up the convergence process, we choose the base model obtained from pre-training on the COCO dataset \citep{lin2014coco} as the starting weight initialization. We also used the mosaic data augmentation method \citep{bochkovskiy2020yolov4} to expand the data. During the testing phase, the batch size of all methods was set to 32, and we utilized FPS to measure prediction speed and did not use acceleration methods such as FP16 or TensorRT to ensure the fairness of the comparison.
\begin{table}[t]
	\centering
	\caption{Experimental results on the DroneVehicle dataset\label{DroneVehicle-contrastive-experiment}}

	\resizebox{\textwidth}{!}{
		\begin{tabular}{lcccccccccc}
			\toprule
			\textbf{Methods}      & \textbf{Modality} & \textbf{Baseline} & \textbf{Car}     & \textbf{Truck}   & \textbf{Bus}     & \textbf{Van}     & \textbf{Freight Car} & \textbf{\(\text{mAP}_{50}\)} (\%) \(\uparrow\) & \textbf{\(\text{mAP}\)} (\%) \(\uparrow \) \\
			\midrule
			RGB-only (YOLOv5s)    & RGB               & YOLOv5            & 91.9             & 56.8             & 72.0             & 93.7             & 58.5                 & 74.6                                           & 46.7                                       \\
			IR-only (YOLOv5s)     & IR                & YOLOv5            & 98.0             & 69.5             & 80.8             & 94.5             & 61.3                 & 80.8                                           & 60.2                                       \\
			RGB-only   (YOLOv8s)  & RGB               & YOLOv8            & 91.9             & 53.1             & 68.8             & 92.7             & 56.5                 & 72.6                                           & 45.7                                       \\
			IR-only (YOLOv8s)     & IR                & YOLOv8            & 98.3             & 69.9             & 76.7             & 97.0             & 66.0                 & 81.6                                           & 61.1                                       \\
			\midrule
			(JSTAR'2021) YOLOrs   & RGB+IR            & YOLOv3            & 97.7             & 77.2             & 96.0             & 65.6             & 63.6                 & 80.0                                           & 58.1                                       \\
			(Arxiv'2022) CFT      & RGB+IR            & YOLOv5            & \underline{98.5} & 75.0             & 82.3             & \underline{97.3} & 68.5                 & 84.3                                           & 61.9                                       \\
			(TGRS'2023) SuperYOLO & RGB+IR            & YOLOv5            & 97.7             & \textbf{79.0}    & 66.3             & 96.6             & 67.6                 & 81.4                                           & 58.6                                       \\
			(TGRS'2023) GHOST     & RGB+IR            & YOLOv5            & 97.3             & 78.8             & 68.9             & 96.3             & 66.5                 & 81.5                                           & 59.3
			\\
			(T-ITS'2023) MFPT     & RGB+IR            & Faster RCNN       & 97.3             & 72.2             & 77.2             & 96.6             & 66.7                 & 82.0                                           & 60.7                                       \\
			(PR'2024) ICAFusion   & RGB+IR            & YOLOv5            & 96.1             & 46.4             & 57.1             & 92.2             & 34.0                 & 65.1                                           & 44.0                                       \\
			(CVPR'2024) GM-DETR   & RGB+IR            & RT-DETR           & 92.4             & 75.3             & 80.8             & 90.8             & 64.9                 & 80.8                                           & 55.9                                       \\
			(CVPR'2024) DaFF      & RGB+IR            & YOLOv5            & 92.2             & 58.9             & 71.9             & 94.4             & 58.2                 & 75.1                                           & 45.5                                       \\
			(TIV'2024) CMADet     & RGB+IR            & YOLOv5            & 98.2             & 70.4             & 78.3             & 96.8             & 66.4                 & 82.0                                           & 59.5                                       \\
			\midrule
			Ours (YOLOv5s)        & RGB+IR            & YOLOv5            & 98.4             & 78.2             & \underline{83.4} & 96.6             & \underline{69.9}     & \underline{85.3}                               & \underline{63.4}                           \\
			Ours (YOLOv8s)        & RGB+IR            & YOLOv8            & \textbf{98.6}    & \underline{78.9} & \textbf{84.1}    & \textbf{97.4}    & \textbf{71.5}        & \textbf{86.1}                                  & \textbf{65.5}
			\\
			\bottomrule
		\end{tabular}
	}
\end{table}
\subsection{Evaluation metrics}
We used the standard mean average precision (mAP) introduced by MS-COCO \citep{lin2014coco} as the primary evaluation metric for the multimodal object detection task. The mAP is calculated as the mean of the average precision (AP) across all classes. The AP is calculated as the area under the precision-recall (P-R) curve, which is obtained by varying the confidence threshold. We used the mAP at an intersection over union (IoU) threshold of 50\% (mAP\(_{50}\)) as the complementary evaluation metric, where mAP\(_{50}\)is calculated by averaging the APs at an IoU threshold of 50\%  across all classes. Bold in the table of experimental results represents the best results and underlining represents the second best results.

\subsection{Experiment1: DroneVehicle Dataset}

\begin{figure*}[t]
	\centering
	\includegraphics[width=0.9\linewidth]{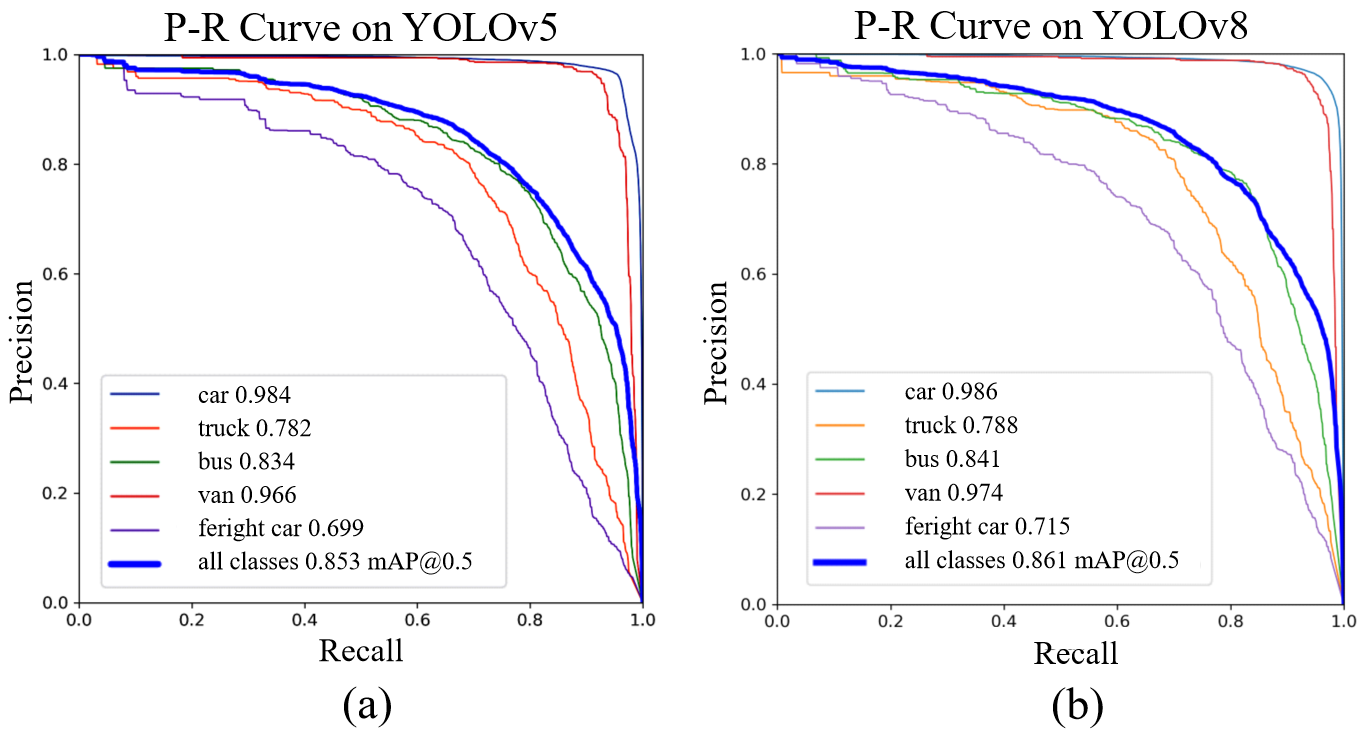}
	\caption{P-R curve results for the COMO method at two different baselines. (a) YOLOv5s baseline. (b) YOLOv8s baseline.}
	\label{fig:pr-curve}
\end{figure*}
\begin{figure*}[t]
	\centering
	\includegraphics[width=0.95\linewidth]{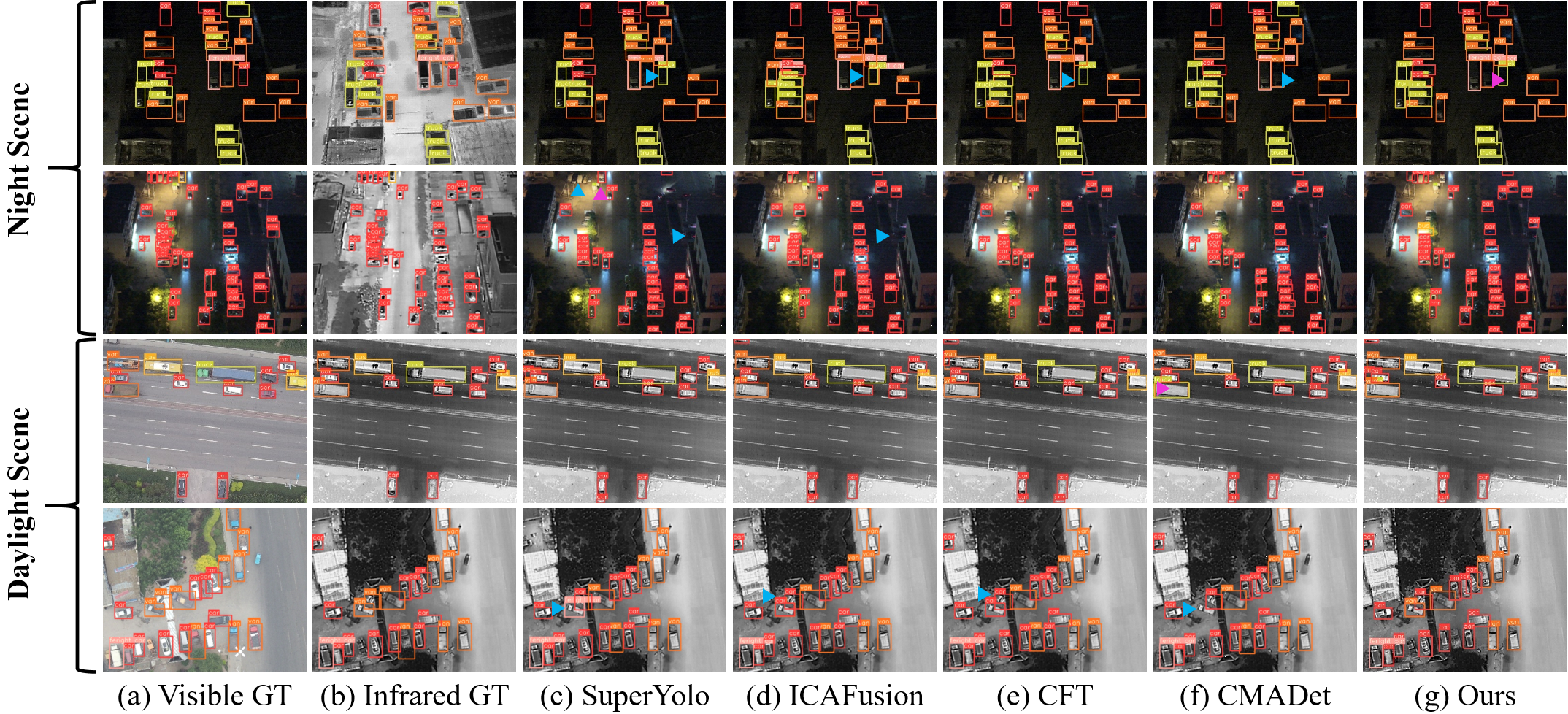}
	\caption{Detection results of the models for different scenarios on the DroneVehicle dataset. We chose to show the detection results of the models on vulnerable modalities to highlight the effectiveness of multimodal fusion. Infrared images are shown for daylight scene and visible images for night scene. The \textcolor{my_blue}{blue} triangles represent missed objects, while the \textcolor{my_magenta}{magenta} triangles indicate incorrectly detected objects.}
	\label{fig:result-DroneVehicle}
\end{figure*}
The DroneVehicle dataset is a large-scale dataset containing images captured by drones in various scenarios, making it highly representative. It provides a total of 28,439 pairs of RGB and infrared images for both day and night scenes. It consists of five categories of targets: car, truck, bus, van, freight car. Due to the positional offset between the two modalities, special consideration is required to achieve optimal detection results. The dataset includes two annotation formats: horizontal and rotated box annotations, with separate labels for each modality. For training, we selected 17,990 image pairs, and for testing, we used 1,469 image pairs. The labeled files from the infrared modality were used as the ground truth for both training and testing.

We compare the results of the proposed method with 9 state-of-the-art methods on the DroneVehicle dataset, as shown in Table \ref{DroneVehicle-contrastive-experiment} and the P-R curve is shown in Fig. \ref{fig:pr-curve}. Our method achieves the best results on both the \(\text{mAP}_{50}\) and \(\text{mAP}\) metrics, with 86.1\% and 65.5\% respectively on the YOLOv8s baseline. Additionally, our method also outperforms other methods on the YOLOv5s baseline, achieving 85.3\% and 63.4\% on the \(\text{mAP}_{50}\) and \(\text{mAP}\) metrics respectively.  These results demonstrate the effectiveness of the COMO approach in multimodal object detection tasks, surpassing existing methods by a significant margin. Furthermore, our method showed notable improvements in detecting large vehicles, such as vans and buses, indicating its capability to make fine distinctions in these cases.

Among the various comparative methods, CFT can achieve the best results because it does not require explicit positional relationships, while other comparative methods rely on explicit positional relationships between modalities, which gives CFT an advantage in cases where the positional relationship between modalities is unclear. Our approach only utilizes high-level features that are less affected by offsets, which can mitigate the impact of offsets on detection results while preserving the information of low-level features. Finally, we use an offset-guided neck fusion network to fuse the features and improve the real-time performance of the model.

Fig. \ref{fig:result-DroneVehicle} illustrates the detection results of our method compared to some other methods on the DroneVehicle dataset. To emphasize the advantages of multimodal object detection, we chose to display visible images from night scenes and infrared images from daylight scenes. This highlights how detection in the weaker modality is enhanced with the support of the other modality. It can be seen that COMO can get the best detection results compared to other methods in complex scenes.

\begin{table}[t]
	\centering
	\caption{Model size, computation cost, and detection speed statistics for different models on the DroneVehicle dataset.}
	\label{tab:params}
    \resizebox{\textwidth}{!}{
	\begin{tabular}{lccc}
		\toprule
		\textbf{Methods} & \textbf{Params}(M) $\downarrow$ & \textbf{Flops}\(@640\)(G) $\downarrow $ & \textbf{FPS}(Hz) $\uparrow $ \\
		\midrule
		(Arxiv'2022) CFT              & 44.76                           & 17.92                                   & 91.74                        \\
		(TGRS'2023) SuperYOLO        & \textbf{4.83}                   & 17.98                                   & 89.4                         \\
		(TGRS'2023) GHOST            & \underline{7.06}                & 20.36                                   & 125.6                        \\
		(T-ITS'2023) MFPT             & 47.65                           & 34.55                                   & 51.2                         \\
		(PR'2024) ICAFusion        & 20.15                           & \underline{14.93}                       & \underline{217.4}            \\
		(CVPR'2024) GM-DETR          & 70.00                           & 176.00                                  & 45.6                         \\
		(CVPR'2024) DaFF             & 45.42                           & 18.45                                   & 85.2                         \\
		(TIV'2024) CMADet           & 33.33                           & 16.86                                   & 208.3                        \\
		\midrule
		Ours (YOLOv5s)   & 16.32                           & \textbf{14.03}                          & 135.1                        \\
		Ours (YOLOv8s)   & 20.27                           & 19.36                                   & \textbf{227.2}               \\
		\bottomrule
	\end{tabular}
 }
\end{table}
At the same time, we compare the model size, computational complexity, and computing speed under the same GPU platform for the compared methods which are denoted by Parameter, Flops, and FPS, respectively. The results are shown in the Table \ref{tab:params}. The results show that our method has a smaller model size and lower computational volume than other methods, and the computing speed is also faster, which indicates that our method has better real-time performance and is more suitable for practical applications. Our method has the smallest computational effort while giving optimal results on the DroneVehicle dataset. And the speed of reasoning is satisfying the display needs.

\subsection{Experiment2: LLVIP Dataset}
\begin{table}[t]
	\centering
	\caption{Contrastive experimental results on the LLVIP dataset\label{tab:LLVIP-contrastive-experiment}}
    \resizebox{\textwidth}{!}{
	\begin{tabular}{lccc}
		\toprule
		\textbf{Methods}   & \textbf{mAP\(_{50}\)} (\%) \(\uparrow\) & \textbf{mAP\(_{75}\)} (\%) \(\uparrow\) & \textbf{mAP} (\%) \(\uparrow\) \\
		\midrule
		RGB-only (YOLOv5s) & 85.9                                    & 62.3                                    & 47.8                           \\
		IR-only (YOLOv5s)  & 94.7                                    & 69.8                                    & 62.5                           \\
		\midrule
		(JSTAR'2021) YOLOrs             & 95.7                                    & 66.3                                    & 61.4                           \\
		(Arxiv'2022) CFT                & 96.5                                    & 68.8                                    & 61.4                           \\
		(TGRS'2023) SuperYOLO          & 93.8                                    & 64.9                                    & 58.1                           \\
		(PR'2024) ICAFusion          & 92.8                                    & 59.1                                    & 47.9                          \\
		(CVPR'2024) GM-DETR            & \underline{97.1}                                    & \textbf{78.8}                                    & \textbf{67.8}                           \\
		(CVPR'2024) DaFF               & 89.1                                    & 51.9                                    & 50.0                             \\
		(TIV'2024) CMADet             & \underline{97.1}                                    & 71.4                                    & 62.9                           \\
		\midrule
		Ours (YOLOv5s)     & \textbf{97.2}                                    & 76.9                                    & \underline{65.3}                           \\
		Ours (YOLOv8s)     & 97.0                                    & \underline{77.1}                                    & 65.2                           \\
		\bottomrule
	\end{tabular}}
\end{table}

\begin{figure}
	\centering
	\includegraphics[width=0.95\linewidth]{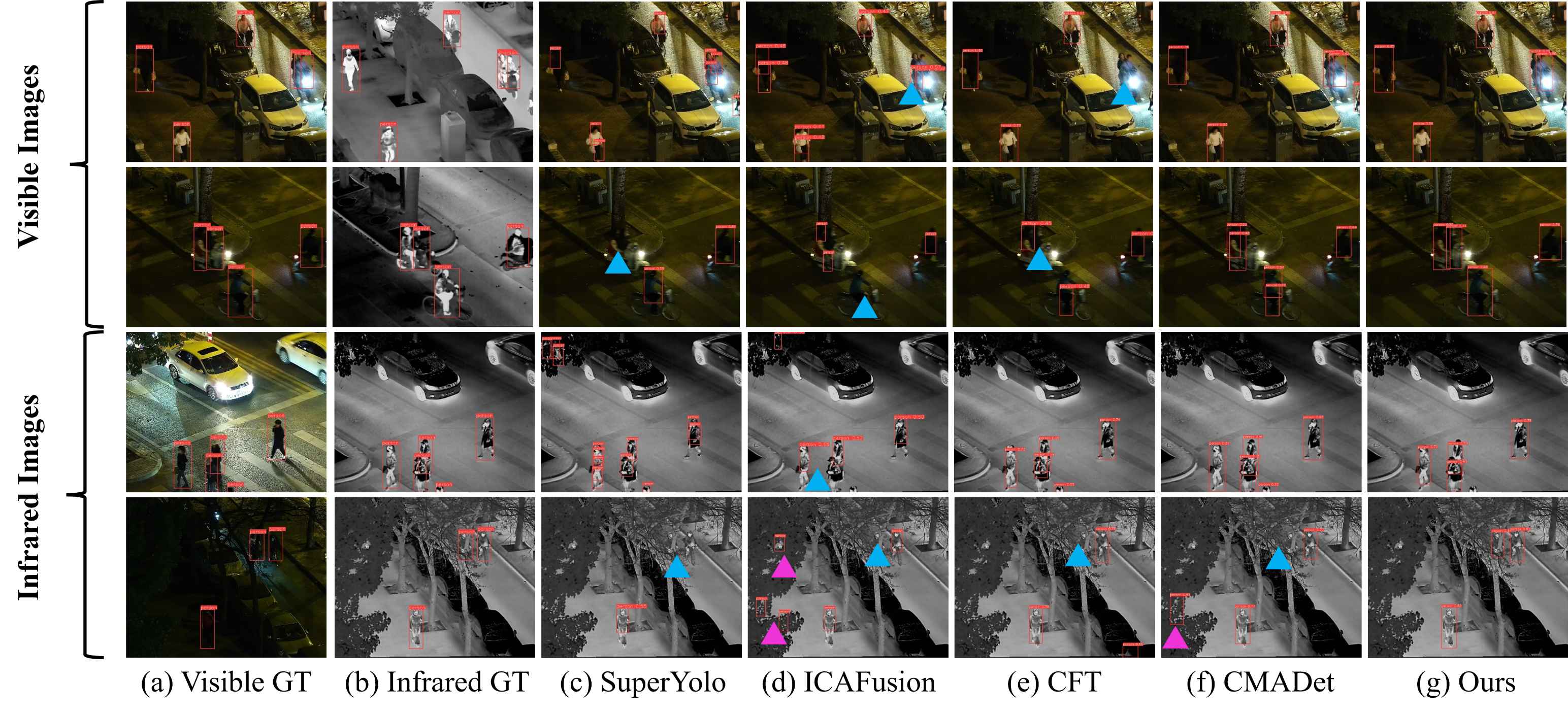}
	\caption{Detection results of the models for different modalities on the LLVIP dataset. The \textcolor{my_blue}{blue} triangles represent missed objects, while the \textcolor{my_magenta}{magenta} triangles indicate incorrectly detected objects.}
	\label{fig:result-llvip}
\end{figure}

Getting good results on multiple datasets is an important way to explore the strengths and weaknesses of a model. Therefore, we chose to use a pedestrian detection dataset similar to the DroneVehicle data perspective, but with only one category. The LLVIP dataset \citep{jia2021llvip} is a challenging dataset containing images in both infrared and visible modalities captured by road surveillance cameras under low light conditions. The dataset includes a total of 16,836 RGB and infrared image pairs.

The LLVIP dataset, with its lower viewing angle and closer proximity to objects, as well as containing only one pedestrian category, presents a slightly lower detection difficulty compared to DroneVehicle. However, the primary challenge with the LLVIP dataset is that it consists entirely of night scenes, making the visible modality significantly less informative. Additionally, the lower viewing angle leads to frequent occlusion of objects by one another. To achieve optimal results on this dataset, it is essential to effectively fuse the infrared data with the visible data while capturing key features of the target. This allows for accurate detection even in cases where occlusion occurs.

We selected 8 comparison methods, and the results are presented in Table \ref{tab:LLVIP-contrastive-experiment}. As the LLVIP dataset contains only one category, we introduce the mAP$_{75}$metric to provide a more comprehensive evaluation of the experimental results, offering additional insights for assessment. Table \ref{tab:LLVIP-contrastive-experiment} demonstrates that our method achieves the best performance on the LLVIP dataset using the YOLOv5 baseline on the mAP\(_{50}\) metric, confirming its effectiveness in multimodal pedestrian detection tasks. Specifically, our method achieved  97.2\% on the mAP\(_{50}\) metric, outperforming the other methods. However, in terms of mAP$_{75}$  and mAP metrics, our method did not surpass the GM-DETR approach, primarily due to its reliance on the RT-DETR \citep{zhao2024rtdetr} baseline, which exhibits higher accuracy for larger targets. In the future, we plan to incorporate a more advanced baseline model to achieve more comprehensive improvements. Additionally, the results obtained using the YOLOv8 baseline are also very close to the best performance, indicating that our method exhibits strong generalization ability and performs well across different datasets.

The qualitative results for the LLVIP dataset are shown in Fig. \ref{fig:result-llvip}. These results demonstrate that our method effectively detects pedestrians in low-light conditions, even when individuals are partially obscured. This highlights the effectiveness of the proposed method in multimodal object detection tasks, particularly in challenging scenarios.

\subsection{Experiment3: VEDAI Dataset}

To further evaluate our proposed method and explore its effectiveness on remote sensing images, we selected the small-scale VEDAI \citep{razakarivony2016vedai} dataset, a widely used benchmark for multimodal remote sensing object detection.
\begin{table*}[t]
	\centering
	\caption{Contrastive experimental results on the VEDAI dataset\label{tab:VEDAI-contrastive-experiment}}

	\resizebox{\textwidth}{!}{
		\begin{tabular}{lccccccccccc}
			\toprule
			\textbf{Methods} & \textbf{Image size} & \textbf{Car} & \textbf{Truck} & \textbf{Pickup} & \textbf{Tractor} & \textbf{Camper} & \textbf{Ship} & \textbf{Van} & \textbf{Plane} & \textbf{mAP\(_{50}\)} (\%) \(\uparrow\) & \textbf{mAP} (\%)\(\uparrow\) \\
			\midrule
			RGB-only         & 512                    & 86.5         & 80.7           & 75.8            & 81.0             & 58.3            & 71.3          & 76.9         & 68.1           & 74.8                                    & 45.2                          \\
			IR-only          & 512                    & 85.3         & 74.9           & 73.3            & 74.9             & 46.8            & 49.2          & 63.0         & \underline{86.2}           & 69.2                                    & 40.1                          \\
			\midrule
			(JSTAR'2021) YOLOrs           & 512                    & 84.2         & 78.3           & 68.8            & 52.6             & 46.8            & 67.9          & 21.3         & 57.9           & 59.7                                    & -                             \\
            (Arxiv'2022) CFT              & 512& 83.2         & 57.5           & 61.0            & 43.6             & 40.8            & 72.9          & 22.3         & 37.3           & 66.6                                    & 38.3                          \\
            (TGRS'2023) GHOST            & 1024                   & 85.6         & 83.1           & 74.9            & 76.9             & 54.4            & 64.3          & 47.7         & 47.9           & 66.9                                    & 40.0                          \\
			(TGRS'2023) superYOLO        & 1024                   & 90.0         & \underline{86.5}           & 71.4            & \underline{83.9}             & 43.3            & 69.7          & \textbf{76.0}         & 83.6           & 76.2                                    & \underline{48.3}                          \\
            (T-ITS'2023) MFPT             & 512                    & 89.5         & 81.0           & \textbf{86.5}            & 71.9             & 53.6            & 70.2          & 64.6         & 74.3           & 74.0                                    & 42.8                          \\
			
			(PR'2024) ICAFusion        & 512                    & \underline{90.1}         & 84.2           & \underline{81.3}            & \textbf{88.8}             & \underline{60.9}            & \underline{73.1}          & \underline{73.6}         & 80.9           & \underline{79.1}                                    & 45.2                          \\
			\midrule
			Ours(YOLOv5s)             & 512                    & \textbf{93.3}         & \textbf{89.7}           & 79.3            & 81.1             & \textbf{62.8}            & \textbf{84.6}          & 73.3         & \textbf{89.7}           & \textbf{81.7}                                    & \textbf{50.3}                          \\
			\bottomrule
		\end{tabular}
	}
\end{table*}

The VEDAI dataset consists of RGB and infrared images captured by aircraft and includes 8 vehicle classes, with over 3,700 annotated targets across more than 1,200 images. The dataset offers images in two resolutions: \(1024\times1024\) and \(512\times512\).  Since this dataset is a well-aligned airborne remote sensing dataset, offset issues are not a concern. Therefore, we applied the feature interaction module to the three feature extraction scales to fuse the data more effectively and obtain richer fusion information. In other words, we utilize the mamba interaction block to perform interaction operations on all input data at each of the three scales \{$S_3^{ir}$,$S_4^{ir}$,$S_5^{ir}$,$S_3^{rgb}$,$S_4^{rgb}$,$S_5^{rgb}$\} 
to obtain the fusion results at the three scales\{$F_3^{ir}$,$F_3^{rgb},F_4^{ir}$,$F_4^{rgb},F_5^{ir}$,$F_5^{rgb}$\}.
\begin{figure}
	\centering
	\includegraphics[width=0.95\linewidth]{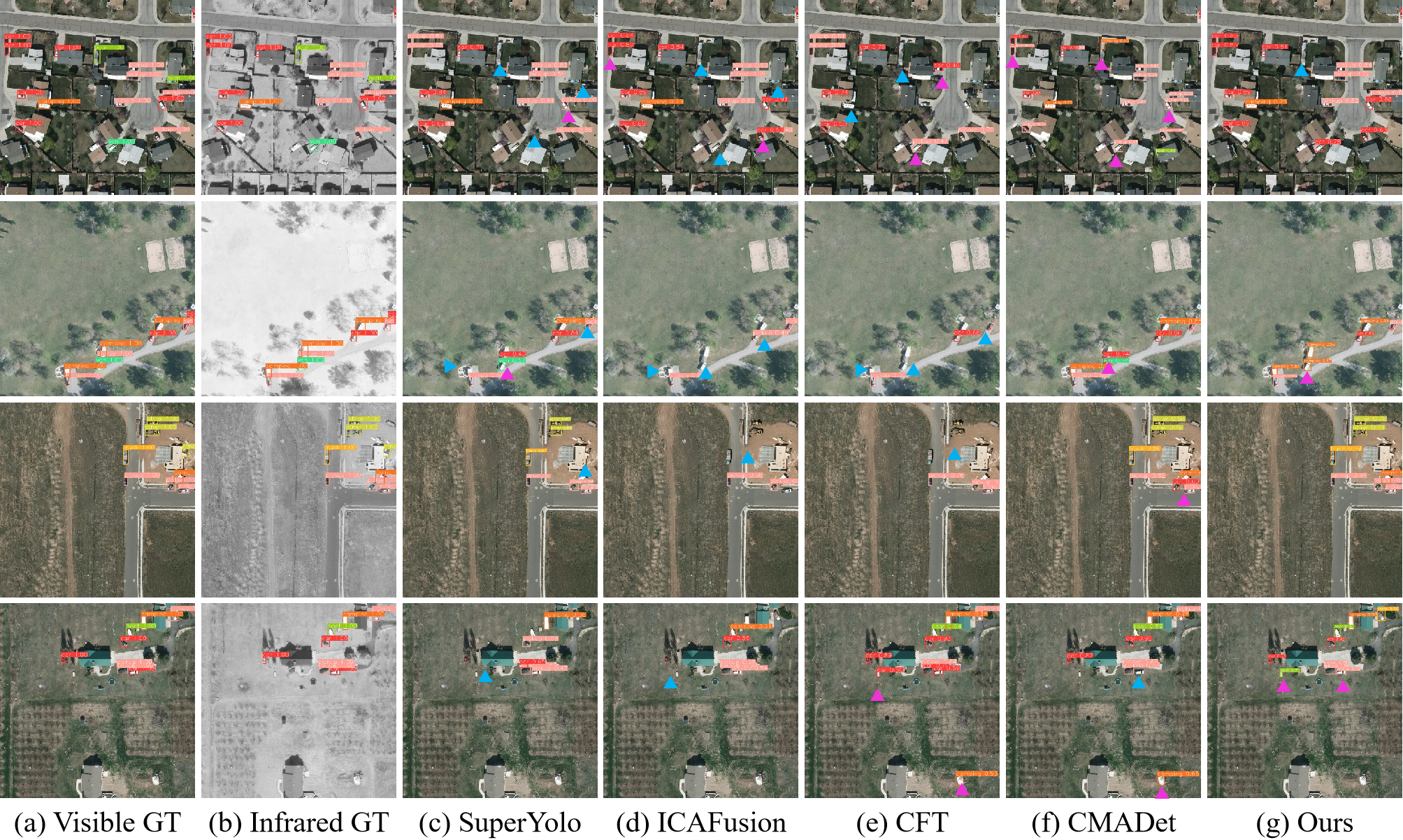}
	\caption{Detection results of the models for different modalities on the VEDAI dataset. The \textcolor{my_blue}{blue} triangles represent missed objects, while the \textcolor{my_magenta}{magenta} triangles indicate incorrectly detected objects.}
	\label{fig:result-vedai}
\end{figure}
For this dataset, we choose only \(512\times512\) resolution images for both training and testing. This choice ensures that the model remains applicable to various datasets, not just the VEDAI dataset in this specific instance. In consideration of other methods that utilized a resolution of \(1024\times1024\), we maintained this setting to ensure that the highest achievable accuracy could be obtained. To expedite the validation of the model's performance, we did not adopt the commonly used ten-fold cross-validation method for the VEDAI dataset. Instead, we fixed one set of data for validation and used the remaining sets for training, reducing the time required for experimentation while still obtaining reliable results. Since all comparison algorithms are based on the same YOLOv5 baseline model, we only compare the experimental results derived from this baseline to ensure a more fair and objective comparison. Additionally, we ensured that the experimental results for all comparison methods were obtained using the same data settings.

The results of the comparison are shown in Table \ref{tab:VEDAI-contrastive-experiment}, which shows that our method also achieves the best results on the VEDAI dataset, demonstrating that our method can also achieve good results on the task of multimodal detection in the remote sensing view. The primary challenge of the VEDAI dataset is that the targets are very small. As a result, when selecting detection heads, only the detection head responsible for detecting small objects can achieve good results. Both SuperYOLO and GHOST adopted this approach. However, this strategy results in the loss of the multiscale fusion network in the neck, which diminishes the effectiveness of multiscale feature fusion.

The results for the VEDAI dataset are shown in Fig. \ref{fig:result-vedai}. These results demonstrate that our method effectively detects small vehicles in remote sensing images, even when the targets are small and the resolution is low. This highlights the effectiveness of the proposed method in multimodal object detection tasks, particularly in challenging remote sensing scenarios.

\subsection{Ablation Study}
We conducted extensive ablation experiments on the proposed modules to explore the effectiveness of each module and the interrelationships between modules. Unless specified mentioned, we primarily utilize the YOLOv5s model as a baseline and the DroneVehicle dataset as experimental data.

As shown in Table \ref{tab:ablation}, we performed numerous ablation experiments to verify the validity of the individual components in the overall model. These include mamba interaction block (MIB), global and local scan method (GLS), and offset-guided fusion (OGF).

Interestingly, the baseline model features a dual-branch architecture comprising two CSPDarknet53 networks for feature extraction. It employs simple convolutional modules for feature fusion before passing the results to the original neck network of YOLOv5 for detection. This baseline design was chosen because the YOLOv5 model alone is insufficient for multimodal object detection tasks. Our network can be seen as an improvement of this baseline, offering improved performance in multimodal object detection tasks. Since the local scan method is an improved method for cross-mamba block, this module cannot be completely isolated for ablation experiments. However, its effect can be explored within the ablation experiments where the cross-mamba block is present.
\begin{table}[t]
	\centering
	\caption{Ablation study with different modules on the DroneVehicle dataset}
	\label{tab:ablation}
	\begin{tabularx}{\textwidth}{Xccccc}
		\toprule
		\textbf{Method}      & \textbf{MIB} & \textbf{GLS} & \textbf{OGF} & \textbf{mAP\(_{50}\)} \(\uparrow\) & \textbf{mAP} \(\uparrow\) \\
		\midrule
		RGB-only             &              &             &              & 74.6                               & 46.7                      \\
		IR-only              &              &             &              & 80.8                               & 60.2                      \\
		\midrule
		(a) Baseline         &              &             &              & 80.9                               & 58.7                      \\
		(b) Baseline+MIB     & \checkmark   &             &              & 83.2                               & 60.5                      \\
		(c) Baseline+OGF     &              &             & \checkmark   & 81.3                               & 58.8                      \\
		(d) Baseline+MIB+LS  & \checkmark   & \checkmark  &              & 83.7                               & 62.1                      \\
		(e) Baseline+MIB+OGF & \checkmark   &             & \checkmark   & 84.1                               & 62.3                      \\
		\midrule
		(f) Ours             & \checkmark   & \checkmark  & \checkmark   & \textbf{85.3}                               & \textbf{63.4}                      \\
		\bottomrule
	\end{tabularx}
\end{table}

The ablation experiments demonstrate that using simple fusion mechanisms like Table    \ref{tab:ablation} (a) yields little improvement in accuracy compared to single-modal object detection tasks. However, the addition of the MIB module (b) significantly improves the detection performance of 2.4\% mAP\(_{50}\) and 0.3 \% mAP, demonstrating the importance of capturing the interaction between different modalities. The local scan method (d) further enhances the fusion of multimodal data with another 0.5\%  mAP\(_{50}\) and 1.6\% mAP, leading to improved detection performance with the local features. In contrast, methods (c) that lack feature interaction and rely solely on multiscale feature fusion show only marginal improvement in performance about 0.5 \% mAP\(_{50}\).

The offset-guided fusion in model (e) further improves the detection performance by 0.9\% mAP\(_{50}\) and 1.8\% mAP compared to model (b), demonstrating that in the presence of the feature interaction method built using the mamba model, guiding low-level features through high-level features, via the mamba-based feature interaction method, significantly mitigates the impact of offset. Finally, the complete model (f) achieves the best performance, with 85.3\% mAP\(_{50}\) and 63.4\% mAP, highlighting the effectiveness of the proposed method in multimodal object detection tasks. This also confirms the necessity of each module and the importance of their rational integration.

\subsection{Comparison and analysis of the mamba interaction block}
\begin{figure}[t]
	\centering
	\includegraphics[width=0.4\linewidth]{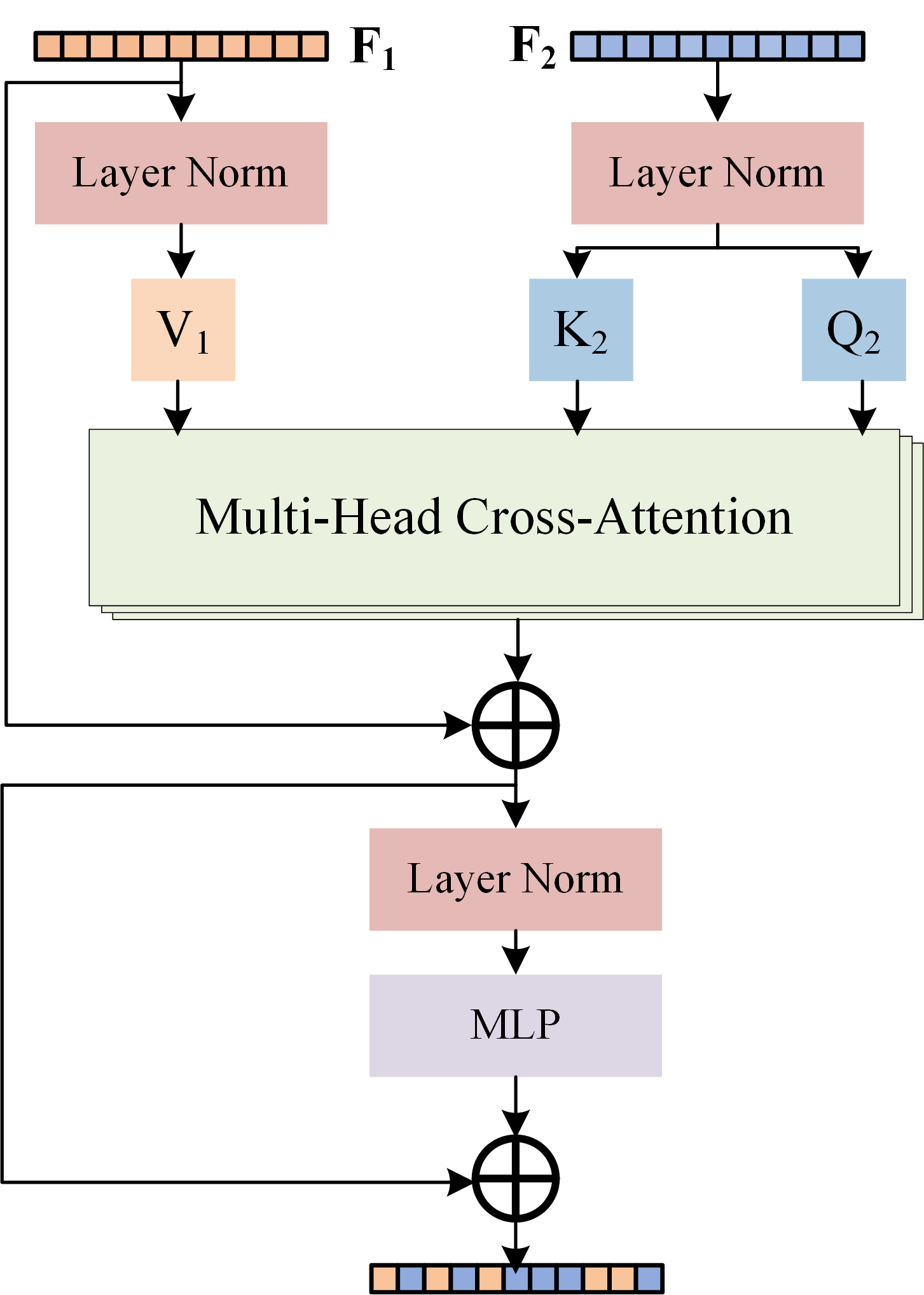}
	\caption{Cross-attention block}
	\label{fig:cross-attention}
\end{figure}

\begin{figure}[t]
    \centering
    \includegraphics[width=0.98\linewidth]{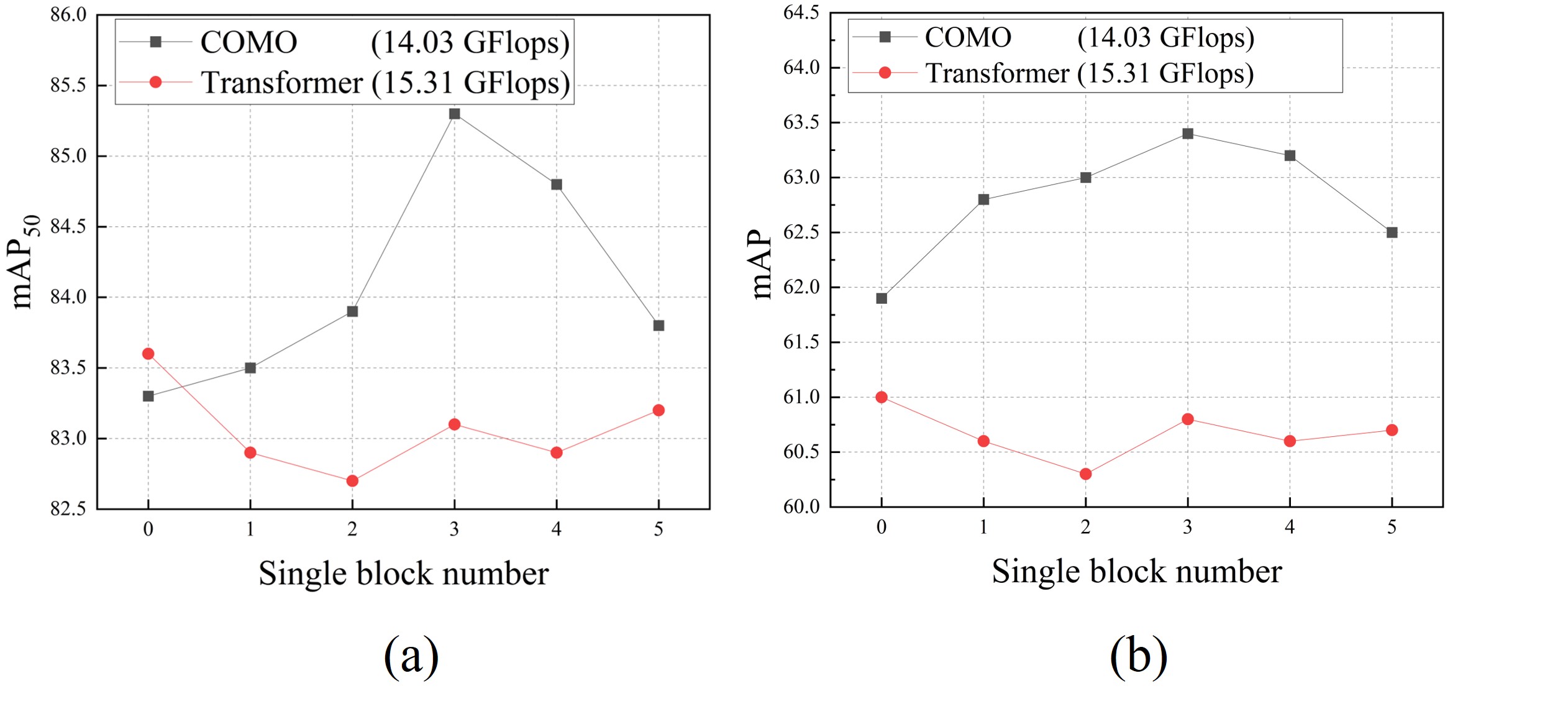}
    \caption{Mamba and Transformer structure analysis and model comparison. (a) The results of mAP$_{50}$. (b) The results of mAP.}
    \label{fig:mamba-transformer}
\end{figure}

In order to find the optimal MIB module and to make an in-depth comparison with other methods, we designed different MIB composition methods and constructed a feature interaction module with a similar structure but consisting entirely of transformer modules, with the aim of fully demonstrating the advantages of the MIB module. The specific operation involved replacing the single-mamba block in the MIB with a self-attention block and substituting the cross-mamba block with a cross-attention block  illustrated in Fig. \ref{fig:cross-attention}. We also analyzed the structure of both models to explore the optimal model for multimodal object detection. 
We analyzed the number of single mode blocks processed in the two models consisting of the mamba model and the transformer model and fixed the number of cross-modalities module processing multiple modal data to 1. The final results are shown in Fig. \ref{fig:mamba-transformer}.

The results show that the MIB module outperforms the transformer module in terms of multimodal object detection tasks. The MIB module achieves the best performance when the number of single blocks number is set to 3, with mAP\(_{50}\) of 85.3\%. In contrast, the transformer module achieves the best performance when the number of single blocks is set to 0 , with mAP\(_{50}\) of 83.6\%. Meanwhile, the FLOPs of the proposed COMO method are only 14.03G, significantly lower than the 15.31G required by methods utilizing transformer models. This demonstrates that the MIB module is more effective than the transformer module in multimodal object detection tasks, as it can better capture the interaction between different modalities and improve detection performance. The results also show that the MIB module is more efficient than the transformer module, as it requires less computation while achieving better performance.

\subsection{Discussion of the global and local scan method}
Adjusting the patch size significantly affects the model's performance and effectiveness. A smaller patch size allows the model to capture finer local features and details, which is essential for detecting small objects or subtle changes in the image. However, this comes at the cost of increased computational load, as the model needs to process more patches. Conversely, a larger patch size enables the model to focus on global information, making it better suited for detecting large objects or broader patterns. The trade-off, however, is a potential loss of detail, particularly in scenarios with small objects or complex backgrounds. Additionally, since our method incorporates a local scanning mechanism, the design of the local window size is closely related to the patch size. Therefore, we conducted an analysis of the patch size and local size in the local scan method to explore the optimal parameters for the local scan method. We analyzed the impact of patch size by considering the number of patches into which a 640×640 image is ultimately divided. The larger the patch size, the fewer the number of patches. These two metrics are specifically represented as \textbf{patch num.} (patch number) and \textbf{local num.} (local window number).
\begin{table}[t]
	\centering
	\caption{Analysis of the local scan method}
	\label{tab:local-scan}
	\begin{tabularx}{\textwidth}{Xccccc}
		\toprule
		\multirow{2}{*}{\textbf{Patch num.}} & \multirow{2}{*}{\textbf{Local num.}} & \multicolumn{2}{c}{single block num.=3} & \multicolumn{2}{c}{single block num.=4}                                            \\
		\cmidrule(lr){3-4} \cmidrule(lr){5-6}
		                            &                             & \textbf{mAP\(_{50}\)}                   & \textbf{mAP}                            & \textbf{mAP\(_{50}\)} & \textbf{mAP}     \\
		\midrule
		\(6 \times 6\)              & \(2 \times 2\)              & 82.3                                    & 60.1                                    & 83.2                  & 61.2             \\
		\(8 \times 8\)              & \(2 \times 2\)              & \textbf{85.3}                           & \textbf{63.4}                                    & \underline{84.5}      & \underline{63.0} \\
		\(9 \times 9\)              & \(3 \times 3\)              & 82.4                                    & 61.5                     & 82.6                  & 61.8             \\
		\(10 \times 10\)            & \(2 \times 2\)              & 84.1                                    & 62.1                                    & 81.8                  & 61.2             \\
		\(20 \times 20\)            & \(4 \times 4\)              & 82.9                                    & 61.5                                    & 81.3                  & 60.7             \\
		\bottomrule
	\end{tabularx}
\end{table}
The results are shown in the Table \ref{tab:local-scan}. We set all local windows to be less than one-third the size of the overall windows to ensure that local information remains more relevant. The results show that the COMO method achieves the best performance when the patch numbers are set to \(8\times8\) and the local size is set to \(2\times2\), with a mAP\(_{50}\) of 85.3\%.  This represents that the local scan method can establish stronger correlations between neighboring patches. 

In remote sensing images, the relationships between local objects are often much stronger. For instance, when scanning a crowded urban area, nearby objects share common patterns and characteristics, making their connections crucial for accurate detection. However, a global sequential scanning mechanism may overlook local details. This is where a local scanning mechanism complements the global approach. Focusing on local details helps bridge the gap, allowing the model to capture both broad patterns and intricate connections between nearby objects. This combination enables the model to build a more comprehensive network of relationships, enhancing its accuracy and effectiveness in detecting features within remote sensing imagery.


\subsection{Discussion of the application scenarios}
The proposed method COMO is designed to address the challenges of multimodal object detection tasks in various application scenarios. It can utilize the rich color and texture information from visible images while also leveraging the radiation information from infrared images. By reasonably fusing the two modalities to create a complementary information set, it achieves higher detection accuracy of target objects, even in conditions such as foggy weather, nighttime, and partial occlusion. This advantage holds significant practical value in real-world applications.

Additionally, COMO addresses the issue of target position offset, a challenge that existing methods struggle to overcome. It mitigates the object offset caused by differences in shooting angles and times by selecting high-level features that encapsulate more abstract attributes of the objects and are less affected by the offset for fusion. Furthermore, it employs the advanced cross-mamba method for inter-modal information interaction, enabling more comprehensive information construction. By using high-level features to guide the fusion of low-level features affected by offset, COMO maximizes the amount of information and ensures the ability to detect small objects. With these methods combined, COMO achieves higher precision in multimodal detection compared to other approaches.

We comprehensively explore the applicability of the COMO method from three perspectives: aerial, drone, and road surveillance, covering most scenarios in the field of remote sensing. In numerous experiments, COMO consistently achieves excellent detection results. Moreover, the required computational resources and processing time are relatively low, making it well-suited for practical applications.

\section{Conclusion}
In this paper, we propose a novel CrOss-Mamba interaction and Offset-guided fusion (COMO) approach for multimodal object detection tasks. This approach leverages the complementary strengths of different modalities to enhance detection accuracy while maintaining real-time processing capabilities. Key components of the COMO approach include the Mamba Interaction Block, Global and Local Scan Method, and Offset-Guided Fusion, which work synergistically to improve multimodal data fusion and detection performance. We validate the effectiveness of the COMO method through experiments on three benchmark datasets, demonstrating its state-of-the-art performance. Additionally, the COMO approach requires fewer computational resources and reduced processing time, making it highly suitable for practical applications in diverse scenarios such as aerial, drone, and road surveillance. In future work, we plan to explore the application of the COMO approach in other multimodal object detection tasks and further optimize the model to achieve even better performance.

\section*{Acknowledgement}
This study was jointly supported by the National Natural Science Foundation of China under Grants 62222116 and 62171417.

\bibliographystyle{elsarticle-harv}
\bibliography{main}

\begin{thebibliography}{50}
\expandafter\ifx\csname natexlab\endcsname\relax\def\natexlab#1{#1}\fi
\providecommand{\url}[1]{\texttt{#1}}
\providecommand{\href}[2]{#2}
\providecommand{\path}[1]{#1}
\providecommand{\DOIprefix}{doi:}
\providecommand{\ArXivprefix}{arXiv:}
\providecommand{\URLprefix}{URL: }
\providecommand{\Pubmedprefix}{pmid:}
\providecommand{\doi}[1]{\href{http://dx.doi.org/#1}{\path{#1}}}
\providecommand{\Pubmed}[1]{\href{pmid:#1}{\path{#1}}}
\providecommand{\bibinfo}[2]{#2}
\ifx\xfnm\relax \def\xfnm[#1]{\unskip,\space#1}\fi
\bibitem[{Althoupety et~al.(2024)Althoupety, Wang, Feng and Rekabdar}]{althoupety2024daff}
\bibinfo{author}{Althoupety, A.}, \bibinfo{author}{Wang, L.Y.}, \bibinfo{author}{Feng, W.C.}, \bibinfo{author}{Rekabdar, B.}, \bibinfo{year}{2024}.
\newblock \bibinfo{title}{Daff: Dual attentive feature fusion for multispectral pedestrian detection}, in: \bibinfo{booktitle}{Proceedings of the IEEE/CVF Conference on Computer Vision and Pattern Recognition}, pp. \bibinfo{pages}{2997--3006}.
\bibitem[{Bochkovskiy(2020)}]{bochkovskiy2020yolov4}
\bibinfo{author}{Bochkovskiy, A.}, \bibinfo{year}{2020}.
\newblock \bibinfo{title}{Yolov4: Optimal speed and accuracy of object detection}.
\newblock \bibinfo{journal}{arXiv preprint arXiv:2004.10934} .
\bibitem[{Burt and Adelson(1987)}]{laplacian1987burt}
\bibinfo{author}{Burt, P.J.}, \bibinfo{author}{Adelson, E.H.}, \bibinfo{year}{1987}.
\newblock \bibinfo{title}{The laplacian pyramid as a compact image code}, in: \bibinfo{booktitle}{Readings in computer vision}. \bibinfo{publisher}{Elsevier}, pp. \bibinfo{pages}{671--679}.
\bibitem[{Cao et~al.(2019)Cao, Guan, Huang, Yang, Cao and Qiao}]{cao2019pedestriandetection}
\bibinfo{author}{Cao, Y.}, \bibinfo{author}{Guan, D.}, \bibinfo{author}{Huang, W.}, \bibinfo{author}{Yang, J.}, \bibinfo{author}{Cao, Y.}, \bibinfo{author}{Qiao, Y.}, \bibinfo{year}{2019}.
\newblock \bibinfo{title}{Pedestrian detection with unsupervised multispectral feature learning using deep neural networks}.
\newblock \bibinfo{journal}{information fusion} \bibinfo{volume}{46}, \bibinfo{pages}{206--217}.
\bibitem[{Carion et~al.(2020)Carion, Massa, Synnaeve, Usunier, Kirillov and Zagoruyko}]{DETR2020carion}
\bibinfo{author}{Carion, N.}, \bibinfo{author}{Massa, F.}, \bibinfo{author}{Synnaeve, G.}, \bibinfo{author}{Usunier, N.}, \bibinfo{author}{Kirillov, A.}, \bibinfo{author}{Zagoruyko, S.}, \bibinfo{year}{2020}.
\newblock \bibinfo{title}{End-to-end object detection with transformers}, in: \bibinfo{booktitle}{European conference on computer vision}, \bibinfo{organization}{Springer}. pp. \bibinfo{pages}{213--229}.
\bibitem[{Chen et~al.(2024a)Chen, Qi, Liu, Bin, Fu, Hu and Zhong}]{chen2024weakly}
\bibinfo{author}{Chen, C.}, \bibinfo{author}{Qi, J.}, \bibinfo{author}{Liu, X.}, \bibinfo{author}{Bin, K.}, \bibinfo{author}{Fu, R.}, \bibinfo{author}{Hu, X.}, \bibinfo{author}{Zhong, P.}, \bibinfo{year}{2024}a.
\newblock \bibinfo{title}{Weakly misalignment-free adaptive feature alignment for uavs-based multimodal object detection}, in: \bibinfo{booktitle}{Proceedings of the IEEE/CVF Conference on Computer Vision and Pattern Recognition}, pp. \bibinfo{pages}{26836--26845}.
\bibitem[{Chen et~al.(2024b)Chen, Song, Han, Xia and Yokoya}]{chen2024changemamba}
\bibinfo{author}{Chen, H.}, \bibinfo{author}{Song, J.}, \bibinfo{author}{Han, C.}, \bibinfo{author}{Xia, J.}, \bibinfo{author}{Yokoya, N.}, \bibinfo{year}{2024}b.
\newblock \bibinfo{title}{Changemamba: Remote sensing change detection with spatiotemporal state space model}.
\newblock \bibinfo{journal}{IEEE Transactions on Geoscience and Remote Sensing} \bibinfo{volume}{62}, \bibinfo{pages}{1--20}.
\bibitem[{Deng and Dragotti(2020)}]{deng2020deng}
\bibinfo{author}{Deng, X.}, \bibinfo{author}{Dragotti, P.L.}, \bibinfo{year}{2020}.
\newblock \bibinfo{title}{Deep convolutional neural network for multi-modal image restoration and fusion}.
\newblock \bibinfo{journal}{IEEE transactions on pattern analysis and machine intelligence} \bibinfo{volume}{43}, \bibinfo{pages}{3333--3348}.
\bibitem[{Dosovitskiy(2020)}]{dosovitskiy2020vit}
\bibinfo{author}{Dosovitskiy, A.}, \bibinfo{year}{2020}.
\newblock \bibinfo{title}{An image is worth 16x16 words: Transformers for image recognition at scale}.
\newblock \bibinfo{journal}{arXiv preprint arXiv:2010.11929} .
\bibitem[{Goodfellow et~al.(2014)Goodfellow, Pouget-Abadie, Mirza, Xu, Warde-Farley, Ozair, Courville and Bengio}]{goodfellow2014gan}
\bibinfo{author}{Goodfellow, I.}, \bibinfo{author}{Pouget-Abadie, J.}, \bibinfo{author}{Mirza, M.}, \bibinfo{author}{Xu, B.}, \bibinfo{author}{Warde-Farley, D.}, \bibinfo{author}{Ozair, S.}, \bibinfo{author}{Courville, A.}, \bibinfo{author}{Bengio, Y.}, \bibinfo{year}{2014}.
\newblock \bibinfo{title}{Generative adversarial nets}.
\newblock \bibinfo{journal}{Advances in neural information processing systems} \bibinfo{volume}{27}.
\bibitem[{Gu and Dao(2023)}]{gu2023mamba}
\bibinfo{author}{Gu, A.}, \bibinfo{author}{Dao, T.}, \bibinfo{year}{2023}.
\newblock \bibinfo{title}{Mamba: Linear-time sequence modeling with selective state spaces}.
\newblock \bibinfo{journal}{arXiv preprint arXiv:2312.00752} .
\bibitem[{Guan et~al.(2019)Guan, Cao, Yang, Cao and Yang}]{guan2019fusion}
\bibinfo{author}{Guan, D.}, \bibinfo{author}{Cao, Y.}, \bibinfo{author}{Yang, J.}, \bibinfo{author}{Cao, Y.}, \bibinfo{author}{Yang, M.Y.}, \bibinfo{year}{2019}.
\newblock \bibinfo{title}{Fusion of multispectral data through illumination-aware deep neural networks for pedestrian detection}.
\newblock \bibinfo{journal}{Information Fusion} \bibinfo{volume}{50}, \bibinfo{pages}{148--157}.
\bibitem[{Guo et~al.(2020)Guo, Fan, Zhang, Xiang and Pan}]{guo2020augfpn}
\bibinfo{author}{Guo, C.}, \bibinfo{author}{Fan, B.}, \bibinfo{author}{Zhang, Q.}, \bibinfo{author}{Xiang, S.}, \bibinfo{author}{Pan, C.}, \bibinfo{year}{2020}.
\newblock \bibinfo{title}{Augfpn: Improving multi-scale feature learning for object detection}, in: \bibinfo{booktitle}{Proceedings of the IEEE/CVF conference on computer vision and pattern recognition}, pp. \bibinfo{pages}{12595--12604}.
\bibitem[{Huang et~al.(2024)Huang, Pei, You, Wang, Qian and Xu}]{huang2024localmamba}
\bibinfo{author}{Huang, T.}, \bibinfo{author}{Pei, X.}, \bibinfo{author}{You, S.}, \bibinfo{author}{Wang, F.}, \bibinfo{author}{Qian, C.}, \bibinfo{author}{Xu, C.}, \bibinfo{year}{2024}.
\newblock \bibinfo{title}{Localmamba: Visual state space model with windowed selective scan}.
\newblock \bibinfo{journal}{arXiv preprint arXiv:2403.09338} .
\bibitem[{Jia et~al.(2021)Jia, Zhu, Li, Tang and Zhou}]{jia2021llvip}
\bibinfo{author}{Jia, X.}, \bibinfo{author}{Zhu, C.}, \bibinfo{author}{Li, M.}, \bibinfo{author}{Tang, W.}, \bibinfo{author}{Zhou, W.}, \bibinfo{year}{2021}.
\newblock \bibinfo{title}{Llvip: A visible-infrared paired dataset for low-light vision}, in: \bibinfo{booktitle}{Proceedings of the IEEE/CVF international conference on computer vision}, pp. \bibinfo{pages}{3496--3504}.
\bibitem[{Li et~al.(2019)Li, Song, Tong and Tang}]{li2019illumination}
\bibinfo{author}{Li, C.}, \bibinfo{author}{Song, D.}, \bibinfo{author}{Tong, R.}, \bibinfo{author}{Tang, M.}, \bibinfo{year}{2019}.
\newblock \bibinfo{title}{Illumination-aware faster r-cnn for robust multispectral pedestrian detection}.
\newblock \bibinfo{journal}{Pattern Recognition} \bibinfo{volume}{85}, \bibinfo{pages}{161--171}.
\bibitem[{Li et~al.(2018)Li, Xiong, An and Wang}]{li2018pan}
\bibinfo{author}{Li, H.}, \bibinfo{author}{Xiong, P.}, \bibinfo{author}{An, J.}, \bibinfo{author}{Wang, L.}, \bibinfo{year}{2018}.
\newblock \bibinfo{title}{Pyramid attention network for semantic segmentation}.
\newblock \bibinfo{journal}{arXiv preprint arXiv:1805.10180} .
\bibitem[{Li et~al.(2022)Li, Hong, Gao, Yao, Zheng, Zhang and Chanussot}]{li2022deep}
\bibinfo{author}{Li, J.}, \bibinfo{author}{Hong, D.}, \bibinfo{author}{Gao, L.}, \bibinfo{author}{Yao, J.}, \bibinfo{author}{Zheng, K.}, \bibinfo{author}{Zhang, B.}, \bibinfo{author}{Chanussot, J.}, \bibinfo{year}{2022}.
\newblock \bibinfo{title}{Deep learning in multimodal remote sensing data fusion: A comprehensive review}.
\newblock \bibinfo{journal}{International Journal of Applied Earth Observation and Geoinformation} \bibinfo{volume}{112}, \bibinfo{pages}{102926}.
\bibitem[{Li et~al.(2017)Li, Kang, Fang, Hu and Yin}]{Pixel2017li}
\bibinfo{author}{Li, S.}, \bibinfo{author}{Kang, X.}, \bibinfo{author}{Fang, L.}, \bibinfo{author}{Hu, J.}, \bibinfo{author}{Yin, H.}, \bibinfo{year}{2017}.
\newblock \bibinfo{title}{Pixel-level image fusion: A survey of the state of the art}.
\newblock \bibinfo{journal}{information Fusion} \bibinfo{volume}{33}, \bibinfo{pages}{100--112}.
\bibitem[{Lin et~al.(2017)Lin, Dollár, Girshick, He, Hariharan and Belongie}]{lin2017FPN}
\bibinfo{author}{Lin, T.Y.}, \bibinfo{author}{Dollár, P.}, \bibinfo{author}{Girshick, R.}, \bibinfo{author}{He, K.}, \bibinfo{author}{Hariharan, B.}, \bibinfo{author}{Belongie, S.}, \bibinfo{year}{2017}.
\newblock \bibinfo{title}{Feature pyramid networks for object detection}.
\newblock \href{http://arxiv.org/abs/1612.03144}{{\tt arXiv:1612.03144}}.
\bibitem[{Lin et~al.(2014)Lin, Maire, Belongie, Hays, Perona, Ramanan, Doll{\'a}r and Zitnick}]{lin2014coco}
\bibinfo{author}{Lin, T.Y.}, \bibinfo{author}{Maire, M.}, \bibinfo{author}{Belongie, S.}, \bibinfo{author}{Hays, J.}, \bibinfo{author}{Perona, P.}, \bibinfo{author}{Ramanan, D.}, \bibinfo{author}{Doll{\'a}r, P.}, \bibinfo{author}{Zitnick, C.L.}, \bibinfo{year}{2014}.
\newblock \bibinfo{title}{Microsoft coco: Common objects in context}, in: \bibinfo{booktitle}{Computer Vision--ECCV 2014: 13th European Conference, Zurich, Switzerland, September 6-12, 2014, Proceedings, Part V 13}, \bibinfo{organization}{Springer}. pp. \bibinfo{pages}{740--755}.
\bibitem[{Liu et~al.(2024a)Liu, Dong, Zhang and Li}]{CRmixing2024liu}
\bibinfo{author}{Liu, C.}, \bibinfo{author}{Dong, Y.}, \bibinfo{author}{Zhang, Y.}, \bibinfo{author}{Li, X.}, \bibinfo{year}{2024}a.
\newblock \bibinfo{title}{Confidence-driven region mixing for optical remote sensing domain adaptation object detection}.
\newblock \bibinfo{journal}{IEEE Transactions on Geoscience and Remote Sensing} .
\bibitem[{Liu et~al.(2016)Liu, Chen, Ward and Wang}]{image2016liu}
\bibinfo{author}{Liu, Y.}, \bibinfo{author}{Chen, X.}, \bibinfo{author}{Ward, R.K.}, \bibinfo{author}{Wang, Z.J.}, \bibinfo{year}{2016}.
\newblock \bibinfo{title}{Image fusion with convolutional sparse representation}.
\newblock \bibinfo{journal}{IEEE signal processing letters} \bibinfo{volume}{23}, \bibinfo{pages}{1882--1886}.
\bibitem[{Liu et~al.(2024b)Liu, Tian, Zhao, Yu, Xie, Wang, Ye and Liu}]{liu2024vmamba}
\bibinfo{author}{Liu, Y.}, \bibinfo{author}{Tian, Y.}, \bibinfo{author}{Zhao, Y.}, \bibinfo{author}{Yu, H.}, \bibinfo{author}{Xie, L.}, \bibinfo{author}{Wang, Y.}, \bibinfo{author}{Ye, Q.}, \bibinfo{author}{Liu, Y.}, \bibinfo{year}{2024}b.
\newblock \bibinfo{title}{Vmamba: Visual state space model}.
\newblock \href{http://arxiv.org/abs/2401.10166}{{\tt arXiv:2401.10166}}.
\bibitem[{Ma et~al.(2024a)Ma, Li and Wang}]{ma2024umamba}
\bibinfo{author}{Ma, J.}, \bibinfo{author}{Li, F.}, \bibinfo{author}{Wang, B.}, \bibinfo{year}{2024}a.
\newblock \bibinfo{title}{U-mamba: Enhancing long-range dependency for biomedical image segmentation}.
\newblock \bibinfo{journal}{arXiv preprint arXiv:2401.04722} .
\bibitem[{Ma et~al.(2024b)Ma, Chai, Jin and Yan}]{ma2024singlemodal}
\bibinfo{author}{Ma, Y.}, \bibinfo{author}{Chai, L.}, \bibinfo{author}{Jin, L.}, \bibinfo{author}{Yan, J.}, \bibinfo{year}{2024}b.
\newblock \bibinfo{title}{Hierarchical alignment network for domain adaptive object detection in aerial images}.
\newblock \bibinfo{journal}{ISPRS Journal of Photogrammetry and Remote Sensing} \bibinfo{volume}{208}, \bibinfo{pages}{39--52}.
\bibitem[{Meher et~al.(2019)Meher, Agrawal, Panda and Abraham}]{meher2019spatialfusion}
\bibinfo{author}{Meher, B.}, \bibinfo{author}{Agrawal, S.}, \bibinfo{author}{Panda, R.}, \bibinfo{author}{Abraham, A.}, \bibinfo{year}{2019}.
\newblock \bibinfo{title}{A survey on region based image fusion methods}.
\newblock \bibinfo{journal}{Information Fusion} \bibinfo{volume}{48}, \bibinfo{pages}{119--132}.
\bibitem[{Nejati et~al.(2015)Nejati, Samavi and Shirani}]{nejati2015multi}
\bibinfo{author}{Nejati, M.}, \bibinfo{author}{Samavi, S.}, \bibinfo{author}{Shirani, S.}, \bibinfo{year}{2015}.
\newblock \bibinfo{title}{Multi-focus image fusion using dictionary-based sparse representation}.
\newblock \bibinfo{journal}{Information Fusion} \bibinfo{volume}{25}, \bibinfo{pages}{72--84}.
\bibitem[{Qingyun et~al.(2021)Qingyun, Dapeng and Zhaokui}]{CFT2021qing}
\bibinfo{author}{Qingyun, F.}, \bibinfo{author}{Dapeng, H.}, \bibinfo{author}{Zhaokui, W.}, \bibinfo{year}{2021}.
\newblock \bibinfo{title}{Cross-modality fusion transformer for multispectral object detection}.
\newblock \bibinfo{journal}{arXiv preprint arXiv:2111.00273} .
\bibitem[{Rao et~al.(2023)Rao, Xu and Wu}]{rao2023tgfuse}
\bibinfo{author}{Rao, D.}, \bibinfo{author}{Xu, T.}, \bibinfo{author}{Wu, X.J.}, \bibinfo{year}{2023}.
\newblock \bibinfo{title}{Tgfuse: An infrared and visible image fusion approach based on transformer and generative adversarial network}.
\newblock \bibinfo{journal}{IEEE Transactions on Image Processing} .
\bibitem[{Razakarivony and Jurie(2016)}]{razakarivony2016vedai}
\bibinfo{author}{Razakarivony, S.}, \bibinfo{author}{Jurie, F.}, \bibinfo{year}{2016}.
\newblock \bibinfo{title}{Vehicle detection in aerial imagery: A small target detection benchmark}.
\newblock \bibinfo{journal}{Journal of Visual Communication and Image Representation} \bibinfo{volume}{34}, \bibinfo{pages}{187--203}.
\bibitem[{Redmon et~al.(2016)Redmon, Divvala, Girshick and Farhadi}]{yolo2016Redmon}
\bibinfo{author}{Redmon, J.}, \bibinfo{author}{Divvala, S.}, \bibinfo{author}{Girshick, R.}, \bibinfo{author}{Farhadi, A.}, \bibinfo{year}{2016}.
\newblock \bibinfo{title}{You only look once: Unified, real-time object detection}, in: \bibinfo{booktitle}{Proceedings of the IEEE conference on computer vision and pattern recognition}, pp. \bibinfo{pages}{779--788}.
\bibitem[{Schreier and Scharf(2010)}]{schreier2010zeroorder}
\bibinfo{author}{Schreier, P.J.}, \bibinfo{author}{Scharf, L.L.}, \bibinfo{year}{2010}.
\newblock \bibinfo{title}{Statistical signal processing of complex-valued data: the theory of improper and noncircular signals}.
\newblock \bibinfo{publisher}{Cambridge university press}.
\bibitem[{Sharma et~al.(2020)Sharma, Dhanaraj, Karnam, Chachlakis, Ptucha, Markopoulos and Saber}]{yolors2020Sharma}
\bibinfo{author}{Sharma, M.}, \bibinfo{author}{Dhanaraj, M.}, \bibinfo{author}{Karnam, S.}, \bibinfo{author}{Chachlakis, D.G.}, \bibinfo{author}{Ptucha, R.}, \bibinfo{author}{Markopoulos, P.P.}, \bibinfo{author}{Saber, E.}, \bibinfo{year}{2020}.
\newblock \bibinfo{title}{Yolors: Object detection in multimodal remote sensing imagery}.
\newblock \bibinfo{journal}{IEEE Journal of Selected Topics in Applied Earth Observations and Remote Sensing} \bibinfo{volume}{14}, \bibinfo{pages}{1497--1508}.
\bibitem[{Shen et~al.(2024)Shen, Chen, Liu, Zuo, Fan and Yang}]{shen2024icafusion}
\bibinfo{author}{Shen, J.}, \bibinfo{author}{Chen, Y.}, \bibinfo{author}{Liu, Y.}, \bibinfo{author}{Zuo, X.}, \bibinfo{author}{Fan, H.}, \bibinfo{author}{Yang, W.}, \bibinfo{year}{2024}.
\newblock \bibinfo{title}{Icafusion: Iterative cross-attention guided feature fusion for multispectral object detection}.
\newblock \bibinfo{journal}{Pattern Recognition} \bibinfo{volume}{145}, \bibinfo{pages}{109913}.
\bibitem[{Song et~al.(2024)Song, Xue, Wen, Ji, Yan and Meng}]{song2024CMA}
\bibinfo{author}{Song, K.}, \bibinfo{author}{Xue, X.}, \bibinfo{author}{Wen, H.}, \bibinfo{author}{Ji, Y.}, \bibinfo{author}{Yan, Y.}, \bibinfo{author}{Meng, Q.}, \bibinfo{year}{2024}.
\newblock \bibinfo{title}{Misaligned visible-thermal object detection: A drone-based benchmark and baseline}.
\newblock \bibinfo{journal}{IEEE Transactions on Intelligent Vehicles} .
\bibitem[{Srivastava et~al.(2014)Srivastava, Hinton, Krizhevsky, Sutskever and Salakhutdinov}]{srivastava2014dropout}
\bibinfo{author}{Srivastava, N.}, \bibinfo{author}{Hinton, G.}, \bibinfo{author}{Krizhevsky, A.}, \bibinfo{author}{Sutskever, I.}, \bibinfo{author}{Salakhutdinov, R.}, \bibinfo{year}{2014}.
\newblock \bibinfo{title}{Dropout: a simple way to prevent neural networks from overfitting}.
\newblock \bibinfo{journal}{The journal of machine learning research} \bibinfo{volume}{15}, \bibinfo{pages}{1929--1958}.
\bibitem[{Sun et~al.(2022)Sun, Cao, Zhu and Hu}]{Sun2022DroneVehicle}
\bibinfo{author}{Sun, Y.}, \bibinfo{author}{Cao, B.}, \bibinfo{author}{Zhu, P.}, \bibinfo{author}{Hu, Q.}, \bibinfo{year}{2022}.
\newblock \bibinfo{title}{Drone-based rgb-infrared cross-modality vehicle detection via uncertainty-aware learning}.
\newblock \bibinfo{journal}{IEEE Transactions on Circuits and Systems for Video Technology} \bibinfo{volume}{32}, \bibinfo{pages}{6700--6713}.
\bibitem[{Xiao et~al.(2024)Xiao, Meng, Wu, Xu, He and Li}]{GMdetr2024xiao}
\bibinfo{author}{Xiao, Y.}, \bibinfo{author}{Meng, F.}, \bibinfo{author}{Wu, Q.}, \bibinfo{author}{Xu, L.}, \bibinfo{author}{He, M.}, \bibinfo{author}{Li, H.}, \bibinfo{year}{2024}.
\newblock \bibinfo{title}{Gm-detr: Generalized muiltispectral detection transformer with efficient fusion encoder for visible-infrared detection}, in: \bibinfo{booktitle}{Proceedings of the IEEE/CVF Conference on Computer Vision and Pattern Recognition}, pp. \bibinfo{pages}{5541--5549}.
\bibitem[{Xie et~al.(2024)Xie, Cui, Ieong, Tan, Zhang, Zheng and Yu}]{xie2024fusionmamba}
\bibinfo{author}{Xie, X.}, \bibinfo{author}{Cui, Y.}, \bibinfo{author}{Ieong, C.I.}, \bibinfo{author}{Tan, T.}, \bibinfo{author}{Zhang, X.}, \bibinfo{author}{Zheng, X.}, \bibinfo{author}{Yu, Z.}, \bibinfo{year}{2024}.
\newblock \bibinfo{title}{Fusionmamba: Dynamic feature enhancement for multimodal image fusion with mamba}.
\newblock \bibinfo{journal}{arXiv preprint arXiv:2404.09498} .
\bibitem[{Yin et~al.(2018)Yin, Liu, Liu and Chen}]{yin2018transformfusion}
\bibinfo{author}{Yin, M.}, \bibinfo{author}{Liu, X.}, \bibinfo{author}{Liu, Y.}, \bibinfo{author}{Chen, X.}, \bibinfo{year}{2018}.
\newblock \bibinfo{title}{Medical image fusion with parameter-adaptive pulse coupled neural network in nonsubsampled shearlet transform domain}.
\newblock \bibinfo{journal}{IEEE Transactions on Instrumentation and Measurement} \bibinfo{volume}{68}, \bibinfo{pages}{49--64}.
\bibitem[{Zhang et~al.(2023a)Zhang, Lei, Xie, Fang, Li and Du}]{superyolo2023zhang}
\bibinfo{author}{Zhang, J.}, \bibinfo{author}{Lei, J.}, \bibinfo{author}{Xie, W.}, \bibinfo{author}{Fang, Z.}, \bibinfo{author}{Li, Y.}, \bibinfo{author}{Du, Q.}, \bibinfo{year}{2023}a.
\newblock \bibinfo{title}{Superyolo: Super resolution assisted object detection in multimodal remote sensing imagery}.
\newblock \bibinfo{journal}{IEEE Transactions on Geoscience and Remote Sensing} \bibinfo{volume}{61}, \bibinfo{pages}{1--15}.
\bibitem[{Zhang et~al.(2023b)Zhang, Lei, Xie, Li, Yang and Jia}]{zhang2023ghost}
\bibinfo{author}{Zhang, J.}, \bibinfo{author}{Lei, J.}, \bibinfo{author}{Xie, W.}, \bibinfo{author}{Li, Y.}, \bibinfo{author}{Yang, G.}, \bibinfo{author}{Jia, X.}, \bibinfo{year}{2023}b.
\newblock \bibinfo{title}{Guided hybrid quantization for object detection in remote sensing imagery via one-to-one self-teaching}.
\newblock \bibinfo{journal}{IEEE Transactions on Geoscience and Remote Sensing} \bibinfo{volume}{61}, \bibinfo{pages}{1--15}.
\bibitem[{Zhang et~al.(2019)Zhang, Liu, Zhang, Yang, Qiao, Huang and Hussain}]{zhang2019cross-modality}
\bibinfo{author}{Zhang, L.}, \bibinfo{author}{Liu, Z.}, \bibinfo{author}{Zhang, S.}, \bibinfo{author}{Yang, X.}, \bibinfo{author}{Qiao, H.}, \bibinfo{author}{Huang, K.}, \bibinfo{author}{Hussain, A.}, \bibinfo{year}{2019}.
\newblock \bibinfo{title}{Cross-modality interactive attention network for multispectral pedestrian detection}.
\newblock \bibinfo{journal}{Information Fusion} \bibinfo{volume}{50}, \bibinfo{pages}{20--29}.
\bibitem[{Zhang and Demiris(2023)}]{zhang2023visible}
\bibinfo{author}{Zhang, X.}, \bibinfo{author}{Demiris, Y.}, \bibinfo{year}{2023}.
\newblock \bibinfo{title}{Visible and infrared image fusion using deep learning}.
\newblock \bibinfo{journal}{IEEE Transactions on Pattern Analysis and Machine Intelligence} \bibinfo{volume}{45}, \bibinfo{pages}{10535--10554}.
\bibitem[{Zhao et~al.(2024)Zhao, Lv, Xu, Wei, Wang, Dang, Liu and Chen}]{zhao2024rtdetr}
\bibinfo{author}{Zhao, Y.}, \bibinfo{author}{Lv, W.}, \bibinfo{author}{Xu, S.}, \bibinfo{author}{Wei, J.}, \bibinfo{author}{Wang, G.}, \bibinfo{author}{Dang, Q.}, \bibinfo{author}{Liu, Y.}, \bibinfo{author}{Chen, J.}, \bibinfo{year}{2024}.
\newblock \bibinfo{title}{Detrs beat yolos on real-time object detection}, in: \bibinfo{booktitle}{Proceedings of the IEEE/CVF Conference on Computer Vision and Pattern Recognition}, pp. \bibinfo{pages}{16965--16974}.
\bibitem[{Zhou et~al.(2020)Zhou, Chen and Cao}]{zhou2020improving}
\bibinfo{author}{Zhou, K.}, \bibinfo{author}{Chen, L.}, \bibinfo{author}{Cao, X.}, \bibinfo{year}{2020}.
\newblock \bibinfo{title}{Improving multispectral pedestrian detection by addressing modality imbalance problems}, in: \bibinfo{booktitle}{Computer Vision--ECCV 2020: 16th European Conference, Glasgow, UK, August 23--28, 2020, Proceedings, Part XVIII 16}, \bibinfo{organization}{Springer}. pp. \bibinfo{pages}{787--803}.
\bibitem[{Zhu et~al.(2024)Zhu, Liao, Zhang, Wang, Liu and Wang}]{zhu2024vim}
\bibinfo{author}{Zhu, L.}, \bibinfo{author}{Liao, B.}, \bibinfo{author}{Zhang, Q.}, \bibinfo{author}{Wang, X.}, \bibinfo{author}{Liu, W.}, \bibinfo{author}{Wang, X.}, \bibinfo{year}{2024}.
\newblock \bibinfo{title}{Vision mamba: Efficient visual representation learning with bidirectional state space model}.
\newblock \bibinfo{journal}{arXiv preprint arXiv:2401.09417} .
\bibitem[{Zhu et~al.(2023)Zhu, Sun, Wang and Huang}]{zhu2023MFPT}
\bibinfo{author}{Zhu, Y.}, \bibinfo{author}{Sun, X.}, \bibinfo{author}{Wang, M.}, \bibinfo{author}{Huang, H.}, \bibinfo{year}{2023}.
\newblock \bibinfo{title}{Multi-modal feature pyramid transformer for rgb-infrared object detection}.
\newblock \bibinfo{journal}{IEEE Transactions on Intelligent Transportation Systems} \bibinfo{volume}{24}, \bibinfo{pages}{9984--9995}.
\bibitem[{Zou et~al.(2023)Zou, Chen, Shi, Guo and Ye}]{zou2023survey}
\bibinfo{author}{Zou, Z.}, \bibinfo{author}{Chen, K.}, \bibinfo{author}{Shi, Z.}, \bibinfo{author}{Guo, Y.}, \bibinfo{author}{Ye, J.}, \bibinfo{year}{2023}.
\newblock \bibinfo{title}{Object detection in 20 years: A survey}.
\newblock \bibinfo{journal}{Proceedings of the IEEE} \bibinfo{volume}{111}, \bibinfo{pages}{257--276}.

\end{thebibliography}
\end{document}